%% file: main.tex
\documentclass[research]{fcs}

\input{src/0-Package.tex}

\title{A Probabilistic Generative Model for Tracking Multi-Knowledge Concept Mastery Probability}

\author[1]{Hengyu Liu}
\author*[1]{Tiancheng Zhang}
\author[1]{Fan Li}
\author[2]{Minghe Yu}
\author[1]{Ge Yu}
\address[1]{School of Computer Science and Engineering, Northeastern University, Shenyang, Liaoning, 110169, China}
\address[2]{Software College, Northeastern University, Shenyang, Liaoning, 110169, China}

\fcssetup{
  received       = {month dd, yyyy},
  accepted       = {month dd, yyyy},
  corr-email     = {tczhang@mail.neu.edu.cn},
}

\begin{abstract}
Knowledge tracing aims to track students' knowledge status over time to predict students' future performance accurately. 
In a real environment, teachers expect knowledge tracing models to provide the interpretable result of knowledge status. 
Markov chain-based knowledge tracking (MCKT) models, such as bayesian knowledge tracing, can track knowledge concept mastery probability over time. However, as the number of tracked knowledge concepts increases, the time complexity of MCKT predicting student performance increases exponentially (also called \textit{explaining away problem}). 
When the number of tracked knowledge concepts is large, we cannot utilize MCKT to track knowledge concept mastery probability over time. 
In addition, the existing MCKT models only consider the relationship between students' knowledge status and problems when modeling students' responses but ignore the relationship between knowledge concepts in the same problem. To address these challenges, we propose an inTerpretable pRobAbilistiC gEnerative moDel (TRACED), which can track students' numerous knowledge concepts mastery probabilities over time. To solve \emph{explain away problem}, we design Long and Short-Term Memory (LSTM)-based networks to approximate the posterior distribution, predict students' future performance, and propose a heuristic algorithm to train LSTMs and probabilistic graphical model jointly. 
To better model students' exercise responses, we proposed a logarithmic linear model with three interactive strategies, which models students' exercise responses by considering the relationship among students' knowledge status, knowledge concept, and problems.
We conduct experiments with four real-world datasets in three knowledge-driven tasks. The experimental results show that TRACED outperforms existing knowledge tracing methods in predicting students' future performance and can learn the relationship among students, knowledge concepts, and problems from students' exercise sequences. 
We also conduct several case studies. The case studies show that TRACED exhibits excellent interpretability and thus has the potential for personalized automatic feedback in the real-world educational environment.
\end{abstract}

\begin{document}
\input{src/1-Introduction}
\input{src/2-RelatedWork}
\input{src/3-ProblemDefinitionAndStudyOverview}
\input{src/4-IKT}
\input{src/5-Experiment}

\input{src/6-Conclusion}
\bibliography{main}
\bibliographystyle{fcs}
\input{src/7-Author.tex}
\end{document}

%% file: src/0-Package.tex
\usepackage{bm}
\usepackage{booktabs} 
\usepackage{enumerate}
\usepackage{algorithm}
\usepackage{algpseudocode}
\usepackage{xspace}
\usepackage{amsfonts}
\usepackage{subfigure}
\usepackage{times}  
\usepackage{helvet}  
\usepackage{courier}  
\usepackage{url}  
\usepackage{eqnarray}
\usepackage{graphicx}
\usepackage{grffile}
\usepackage{longtable}
\usepackage{wrapfig}
\usepackage{rotating}
\usepackage[normalem]{ulem}
\usepackage{textcomp}
\usepackage{amssymb}
\usepackage{capt-of}
\usepackage{stfloats}
\usepackage{subfigure}
\usepackage{algorithm}
\usepackage{amsmath}
\usepackage{graphicx}
\usepackage{multirow}
\allowdisplaybreaks[1]

\input{concept_list}
\usepackage{algpseudocode}  
\usepackage{amsmath}


%% file: src/1-Introduction.tex
\section{Introduction}\label{sec:org477f836}
\begin{figure}[!htbp]
\centering{
\includegraphics[width=0.45\textwidth]{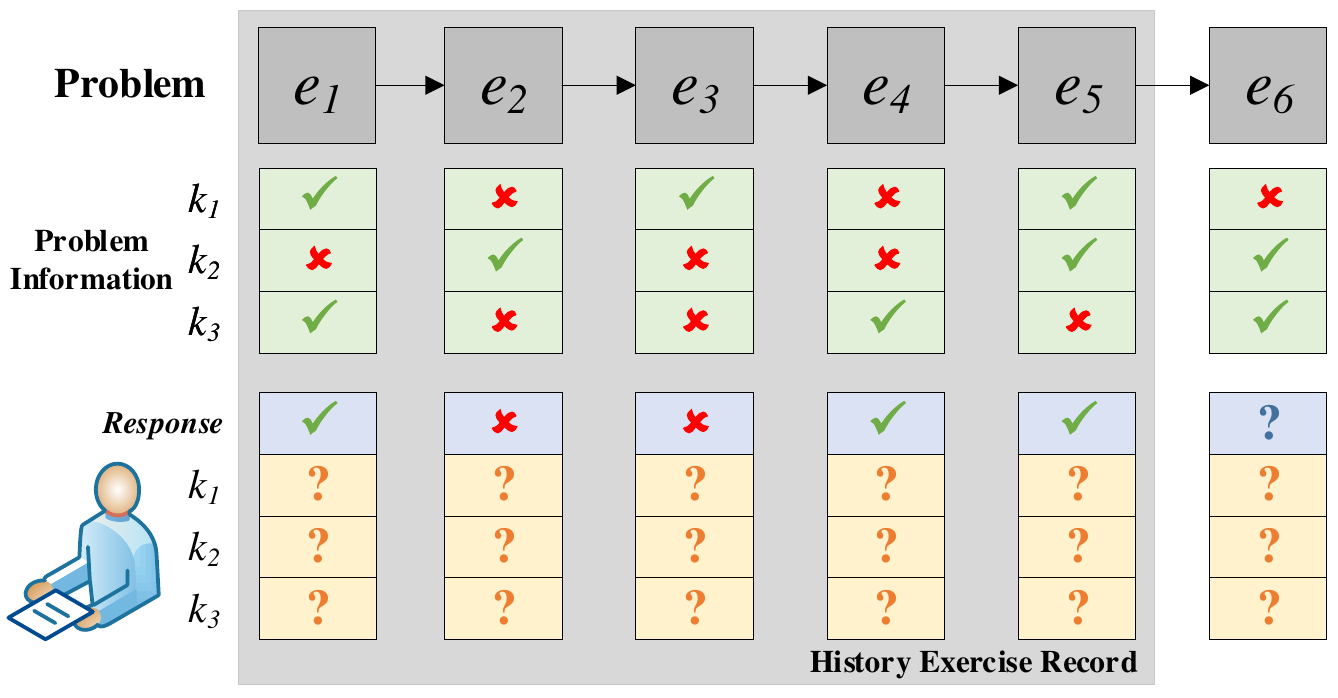}}
\caption{\label{fig:ExampleKnowledgeTracing} A toy example of the knowledge tracking task.}
\end{figure}

Recently, the number of online learning students continually increased with the development of intelligent online education, such as Massive Open Online Courses (MOOCs) \cite{DBLP:conf/kdd/LiuTLZCMW19} and Online Judging (OJ) \cite{DBLP:conf/ijcai/WuLLCSCH15} systems. 
The teacher-student ratio is getting higher and higher.
A teacher may serve thousands of learners simultaneously and cannot provide personalized service for learners. Therefore, there is an increasing demand for recommendation applications based on online intelligent education, including exercise and learning path recommendations \cite{DBLP:conf/edm/AiCGZWFW19}. A key issue in such applications is knowledge tracing, i.e., capturing students' knowledge status over time \cite{DBLP:conf/nips/PiechBHGSGS15}.
As shown in Fig. \ref{fig:ExampleKnowledgeTracing}, knowledge tracing models track the change in students' past knowledge status over time utilizing problem information (knowledge concepts contained in problems), students' historical assignment problems and their responses. On this basis, knowledge tracing models predict students' future knowledge status and learning performance. Such knowledge tracing models can help teachers adjust teaching workloads and future teaching plans.

In a real environment, teachers generally want to obtain interpretable results of students' knowledge status through knowledge tracing models. 
Markov chain-based knowledge tracking (MCKT) models 
\cite{DBLP:conf/ijcai/WuLLCSCH15,DBLP:conf/cikm/ChenLHWCWSH17,Sun2022,Wong2021,Liu2022,Zhang2018} 
can provide interpretable results of students' knowledge status (knowledge concept mastery probability) for teachers. MCKT models regard the observed student's exercise performance and the unobservable student's knowledge status as observed and latent random variables, respectively. Then, MCKT models the relationship between random variables based on prior knowledge and then tracks knowledge concept mastery probability over time. 
However, as the number of knowledge concepts increases, the number of latent variables keeps increasing. These latent variables may all relate to an observed variable (student assignment performance), which exponentially increases the time complexity of MCKT in calculating the student performance probability. 
It is also known as the \textit{explaining away problem} \cite{204911}. 
In addition, the existing MCKT models only consider the relationship between students' knowledge status and their performance when modeling students' exercise response but ignore the relationship between knowledge concepts involved in a learning task. 

To address these challenges, we propose an inTerpretable pRob-
AbilistiC gEnerative moDel (also called as TRACED) for tracking knowledge concept mastery probability over time. 
Reasonable prior is critical to the performance of probabilistic graphical models. 
So, to better model the change of knowledge state over time, we adopt the learning curve and forgetting curve as priors to track students' knowledge concept mastery probabilities over time. 
To better model the students' exercise response (i.e., students' performance), we consider the relationship among students' knowledge status, knowledge concept and problems.
Specifically, we design a logarithmic linear module with three interactive strategies, which model students' exercise responses by considering the relationship among students' knowledge status, knowledge concept and problems. 
Our designed logarithmic linear module can also acquire distributed representation of students' knowledge status, knowledge concepts and problems. 
To solve \textit{explaining away problem} (i.e., the high time complexity of posterior estimation and students' future performance prediction), we utilize two LSTM networks to predict students' future performance and approximate the posterior distribution in model inference, respectively. 
Furthermore, we propose a heuristic algorithm to train probabilistic generative models and LSTMs jointly. 
The joint training of LSTMs and probabilistic generative models is divided into two phases, i.e., the Wake and Sleeping Phases. 
During the Wake Phase, we utilize an LSTM network to approximate the posterior distribution and then sample the latent variables for each training sample based on the LSTM network. Then we learn the parameters of the probabilistic generative model by maximizing the Evidence Lower BOund (ELBO). 
At the Sleep Phase, we first generate new data with completed latent variables by using our probabilistic generative model. 
Then these new data are used to train the LSTM network for approximating the posterior distribution. 
Repeating the above two stages, we jointly train the probability generation model and the LSTM.

To verify the effectiveness of our model, we conduct experiments with four real-world datasets in three knowledge-driven tasks. 
Three knowledge-driven tasks are future student performance prediction, knowledge concept relation prediction, and knowledge concept prediction for problems.
The experimental results in future student performance prediction show that TRACED outperforms other knowledge tracing methods in predicting students' future performance. 
The experimental results in knowledge concept relation prediction and knowledge concept prediction for problems show that our model can learn the relationship among students' knowledge status, knowledge concept and problem from students' exercise sequences. 
Moreover, the distributed representation of knowledge concepts and problems learned by TRACED is effective. 
Further, we perform several case studies to show that TRACED exhibits excellent interpretability and has the potential for personalized automatic feedback in a real-world educational environment. 

The main contributions are outlined as follows:

\begin{itemize}
\item To the best of our knowledge, this is the first comprehensive attempt to track students' numerous knowledge concept mastery probabilities over time instead of tracking knowledge state by modeling the mastery of knowledge concepts as unexplainable model parameters or hidden states. 

\item In order to better model students' exercise responses, we design a logarithmic linear module with three interactive strategies, which model students' exercise responses by considering the relationship among students' knowledge status, knowledge concept and problems. 

\item To solve the \emph{explaining away problem} (i.e., the high time complexity of posterior estimation and students' future performance prediction), we utilize two LSTM-based networks to approximate the posterior distribution and predict students' future performance, respectively. Then we propose a heuristic algorithm to train LSTMs and probabilistic generative model jointly. 

\item Experiments show that TRACED is efficient on four real-world datasets for three knowledge driven tasks (future performance prediction, knowledge concept relation prediction and concept prediction for problems) and exhibits excellent interpretability.  
\end{itemize}

%% file: src/2-RelatedWork.tex
\section{Related Work}
\label{sec:orgb8b682f}

We summarize existing knowledge tracing methods in the following three categories: Factor Analysis-based Knowledge Tracing (FAKT) model, Probabilities Graph-based Knowledge Tracing (PGKT) model and Deep Learning-based Knowledge Tracing (DLKT) model. 

\subsection{Factor Analysis-based Knowledge Tracing} 
FAKT models tend to learn common factors (such as students' ability, problem difficulty, e.g.) in data (students' history exercise) to predict students' performance. 
Moreover, these models consider time factors by counting temporal features (such as the number of wrong attempts and correct attempts, and soon on.). The most simple model of FAKT is the 1-parameter logistic Item Response Theory (IRT) \cite{van2013handbook}, also known as the Rasch model; it does not consider learning among several attempts. The additive factor model (AFM) \cite{cen2006learning} considers the number of attempts a student has made to a problem. The performance factor analysis (PFA) model \cite{pavlik2009performance} separately counts positive and negative attempts as temporal features. Vie and Kashima proposed a family of models named Knowledge Tracing Machines (KTM) \cite{DBLP:conf/aaai/VieK19}, which encompasses IRT, the AFM and PFA as special cases. DAS3H \cite{DBLP:conf/edm/ChoffinPBV19} takes into account both memory decay and multiple knowledge concept tagging by including a representation of the temporal distribution of past exercise on the knowledge concepts involved by a problem. However, FAKT models ignore the order of students' historical exercises. Although FAKT can complement the data with temporal features such as simple counters, FAKT can only get a global students' knowledge status and cannot track knowledge status over time. 

\subsection{Probabilistic Graph-based Knowledge Tracing}
PGKT models regard the observed students' exercise performance and the unobservable students' knowledge status as observed and latent random variables. Then PGKT models the relationship between random variables based on prior knowledge and predicts students' future performance by analyzing and modeling students' exercise process.
Probabilistic graph-based knowledge tracing can be divided into two categories, real-time knowledge tracing (also called Markov chain-based Knowledge Tracing, MCKT) and non-real-time probabilistic graph-based knowledge tracing. 

\noindent \textbf{Markov Chain-based Knowledge Tracing.} The representative of real-time knowledge tracing is Bay-esian Knowledge Tracing (BKT).
BKT \cite{corbett1994knowledge} assumes that a student's knowledge mastery can be represented as a set of binary variables. Each binary variable indicates whether a student has mastered a particular knowledge concept, and a student's knowledge mastery is modeled as a latent variable in a hidden Markov model.
Gorgun et al. \cite{Gorgun2022} analyzed the influence of student disengagement on prediction accuracy in BKT.
Zhang et al.\cite{Zhang2018} proposes a bayesian knowledge tracing model with three learning states. They divide a learning process into three sections by using an evaluation function for three-way decisions.
FBKT \cite{Liu2022} are proposed to address continuous score scenarios (e.g., subjective examinations) so that the applicability of BKT models may be broadened. 

\noindent \textbf{Non-real-time probabilistic graph-based knowledge tracing.} The representative of the non-real-time knowledge tracing is Deterministic Input Noisy-and-gAte (DINA). Although they can diagnose the knowledge states of learners at a certain moment, they need to efficiently track the knowledge states of learners over time and consider factors such as learning and forgetting in the learning process of the learners.
DINA \cite{de2009dina} is a parsimonious and interpretable model that models knowledge mastery as a multidimensional binary latent vector and requires only two parameters (i.e., slipping parameter and guessing parameter) for each problem regardless of the number of knowledge concepts considered.
Fuzzy Cognitive Diagnosis Framework \cite{DBLP:conf/ijcai/WuLLCSCH15} combines fuzzy set theory and educational hypotheses to model a student's knowledge proficiency and predicts a student's performance by considering both the slipping factor and guessing factor.
Knowledge Proficiency Tracing \cite{DBLP:conf/cikm/ChenLHWCWSH17} is an explanatory probabilistic method that tracks the knowledge mastery of students over time by leveraging educational priors (i.e., Q-matrix).
Knowledge Tracing model with Learning Transfer \cite{liu2020tracking} proposed a probabilistic gra-phical model which tracks a student's knowledge proficiency, abstract principle mastery level, and knowledge structure by applying the cognitive structure migration theory  \cite{ausubel1968educational} as priors.
However, these models have strong assumptions since the partition function is difficult to calculate and \emph{explain away problem}. They can not track students' numerous knowledge concept mastery probabilities over time. 

\subsection{Deep Learning-based Knowledge Tracing} 
With the development of deep learning in recent years, LSTM, MANN, and Attention Mechanisms have been applied to the knowledge tracing task. Deep Knowledge Tracing (DKT) \cite{DBLP:conf/nips/PiechBHGSGS15} was the first model to apply deep learning algorithms for knowledge tracing. DKT uses flexible recurrent neural networks that are `deep' in time to track students' knowledge mastery. Subsequently, the Exercise-Enhanced Recurrent Neural Network model \cite{DBLP:conf/aaai/SuLLHYCDWH18} (EERNN) has been proposed based on DKT to take advantage of students' learning records and the text of each problem. Exercise-aware Knowledge Tracing \cite{huang2019ekt} is a framework extended from the EERNN by incorporating knowledge concept information, where the student's integrated state vector is now extended to a knowledge state matrix. However, the knowledge state matrix still needs to be explained.
Lu et al. \cite{lu2020towards} applying the layer-wise relevance propagation method to interpret the RNN-based DLKT model by backpropagating the relevance from the model's output layer to its input layer. 
However, this method can only solve the interpretability of the model, not the interpretability of the tracking results of knowledge concept mastery.
DKVMN \cite{DBLP:conf/www/ZhangSKY17} applies key-value memory networks to exploit the relationships among the underlying knowledge and directly outputs a student's knowledge proficiency. DKVMN-Context Aware \cite{DBLP:conf/edm/AiCGZWFW19} modifies the DKVMN to design its memory structure based on the course's concept list and explicitly considers the knowledge concept mapping relationship during knowledge tracing. 
DKVMN Decision Tree \cite{sun2019muti} improves the performance of the DKVMN model by incorporating additional features to the input, which applies a DT classifier to preprocess the behavior features.
Self-Attentive Knowledge Tracing \cite{DBLP:conf/edm/PandeyK19} is an approach that uses an attention mechanism to identify a problem from the student's past activities that are relevant to the given problem and predicts students' exercise responses.
Relation Aware Knowledge Tracing \cite{DBLP:conf/cikm/PandeyS20} is a relation-aware self-attention model that uses a relation-aware self-attention layer to incorporate contextual information. The contextual information integrates exercise relation information via their textual content as well as students' performance data and forgotten behavior information.
Attentive Knowledge Tracing \cite{ghosh2020context} employs a monotonic attention mechanism, which relates students' future responses to assessment problems to their past responses to predict students' future performance. 
Zhu et al. \cite{zhu2020learning} propose an effective attention-based model for tracking knowledge state, which captures the relationships among each item of the input regardless of the length of the input sequence. 
And Yu et al. \cite{yu2022contextkt} propose a context-based knowledge tracing model, which combines students’ historical performance and their studying contexts during knowledge mastery. 
However, since the internal logic to achieve the desired output or result that is un-understandable and unexplainable, DLKT is less interpretable in tracing students' knowledge status. 
Although there are some works \cite{DBLP:conf/kdd/0001HL20} that attempt to address \textit{black box problems} of DLKT, they explain how DLKT predicts student exercise feedback by showing the impact weight of students' historical exercise records on predicting students' current exercise responses. 

%% file: src/3-ProblemDefinitionAndStudyOverview.tex
\begin{table*}[!ht]
\caption{\textbf{Key Notations in IKT}}
\label{table:somekeynotationsinkcrl}
\centering
\begin{tabular}{l | c | l}
\hline
& notation        & description \\ 
\hline
\multirow{8}{*}[-1mm]{ \centering \rotatebox{90}{\centering  Dataset Description}}
 & N               & the total number of students\\                       
 & M               & the total number of problems\\                          
 & K               & the total number of knowledge concepts\\
 & $S_{i}$         & student $i$'s exercise record\\
 & $r_i$           & the response result sequence of student $i$'s exercise record\\
 & $e_i$           & the problem sequence of student $i$'s exercise record\\
 & $\tau_i$           & the time sequence of student $i$'s exercise record\\
 & $Q_{j,k}$       & problem $j$ contains knowledge concept $k$ or not\\
\hline 
\multirow{11}{*}[-1mm]{\centering \rotatebox{90}{\centering  Model Parameters}}
& $E_{e,j}$       & the distributed representation of problem $j$\\
& $E_{c,k}$       & the distributed representation of knowledge concept $k$\\ 
& $\pi_{k}$       & the probability that students initially master the knowledge concept $k$\\
& $\theta_{s,j}$  & problem $j$'s slipping parameters \\
& $\theta_{g,j}$  & problem $j$'s guessing parameters \\ 
& $\theta_{l,k}$  & knowledge concept $k$'s learning parameters\\
& $\theta_{f,k}$  & knowledge concept $k$'s forgetting parameters\\
& $b_{l,k}$       & knowledge concept $k$'s learning bias\\
& $b_{f,k}$       & knowledge concept $k$'s forgetting bias\\
& $w_{e,j}$       & problem $j$'s bias \\
& $w_{c,k}$       & knowledge concept $k$'s bias\\
& $\mathbf{Z}_{**}, \mathbf{b}_{*}$ & the parameters in LSTM \\
\hline 
\multirow{6}{*}[+2mm]{\centering \rotatebox{90}{\centering  Random Variable}}
& $u_{i,k}^{t}$   & student $i$ masters knowledge concept $k$ or not at the $t$-th exercise record\\
& $s_{j}$         & students made a mistake on problem $j$\\	  & $g_{j}$         & students answer problem $j$ by guessing\\
& $f_{k}$         & students forget knowledge concept $k$\\    
& $l_{k}$         & students master knowledge concept $k$ through learning\\
\hline
\multirow{5}{*}[-3mm]{\centering \rotatebox{90}{\centering  Hyperparameter}}
& &\\
& $d_e$           & the dimension of distributed representation\\
& $d_{h}$       & the dimension of hidden state in the LSTM which approximates posterior distribution\\
& $d_{p}$       & the dimension of hidden state in the LSTM which predicts students' future performance\\
& $\Delta \hat{\tau}$& the time interval for calculating knowledge concepts exercise frequency  \\
& &\\
\hline
\end{tabular}
\end{table*}
\section{Problem Definition and Study \\ Overview}
\label{sec:org55cd7f2}
In this section, we first formally introduce knowledge tracing and three knowledge-driven tasks. Then we introduce our study overview. 

Assume that there are \(N\) students, \(M\) problems, and \(K\) knowledge concepts in a learning system. 
In this system, student exercise logs record the results of exercises performed by the students at different times. 
Student \(i\)'s exercise sequence is represented as \(S_i = \left\{ S_{i,1}, S_{i,2}, S_{i,3}..., \right\}\), and 
\(S_{i,t} = \left\{ e_{i,t}, r_{i,t}, \tau_{i,t} \right\}\) represents that student \(i\) submitted problem \(e_{i,t}\) as part of the \emph{t}-th exercise record. The submission time was \(\tau_{i,t}\), and the result was \(r_{i,t}\) (either ``True" or ``False'', and we code True=1 and False=0). 
Additionally, we have a Q-matrix, which is represented as a binary matrix \(Q \in \mathbb{R}^{M*K}\). \(Q_{j,k}=1\) means that problem \(j\) is related to the knowledge concept \(k\), where a value of 0 indicates that the corresponding problem and concept are unrelated.
Without a loss of generality, knowledge tracing and three knowledge-driven tasks are formulated as follows: 

\noindent \textbf{Knowledge Tracing.} Given students' exercise sequence \(S\), and Q-matrix labeled by educational experts, our goal is three-fold: 1) modeling students' knowledge concepts mastery probabilities from the 1-st to $t$-th exercise; 2) predicting students' knowledge concepts mastery probabilities at the ($t+1$)-th exercise; 3) calculating the distributed representation of students' exercise records from the 1-st to ($t+1$)-th exercise, the distributed representation of knowledge concepts, and the distributed representation of problems. 

\noindent \textbf{Future Performance Prediction Task.} 
Given the distributed representations of students' exercise records from the 1st to ($t+1$)-th exercise, our goal is to predict students' responses at the ($t+1$)-th exercise. 

\noindent \textbf{Knowledge Concept Relation Prediction Task.} Given distributed representations of a set of knowledge concepts, the goal is to predict the relationships among knowledge concepts. The relationships between knowledge concepts include superior relationships, subordinate relationships, and disparate relationships.

\noindent \textbf{Concept Prediction for Problems Task.} Given distributed representations of knowledge concepts and problems, the goal is to predict the knowledge concepts of the given problems.

As shown in Figure \ref{fig:IKTFramework}, our solution is a two-stage framework, which contains a modeling stage and predicting stage: 1) In the modeling stage, given students' exercises records and Q-matrix, 
we first model students' learning and forgetting behavior to track students' knowledge concept mastery probabilities over time. 
Then, we proposed a logarithmic linear model to model students' exercise responses by analyzing the interactions among students, problems, and knowledge concepts. After that, we can obtain students' knowledge concept probabilities \(U^{1}, ..., U^{t}\) at different times and students' learning trajectories. 2) In predicting stage, TRACED predicts students' future responses \(R^{t+1}\) and future knowledge concept mastery probabilities \(U^{t+1}\) in the future. 

Besides, we also provide the learned distributed representations of knowledge concepts and problems as pre-trained parameters to the fully connected neural network for knowledge concept relation prediction and concept prediction for problems. 

%% file: src/4-IKT.tex
\section{The Interpretable Probabilistic Generative Model}
\label{sec:org556a3dd}
In this section, we introduce the technical details of TRACED and how to apply it to three knowledge-driven tasks. For better illustration, key notations in TRACED, which can be divided into four categories, are summarized in Table \ref{table:somekeynotationsinkcrl}, namely, dataset description, model parameters, random variable, and hyperparameter notation.

\begin{figure}[!tbp]
\centering{
\includegraphics[width=0.48\textwidth]{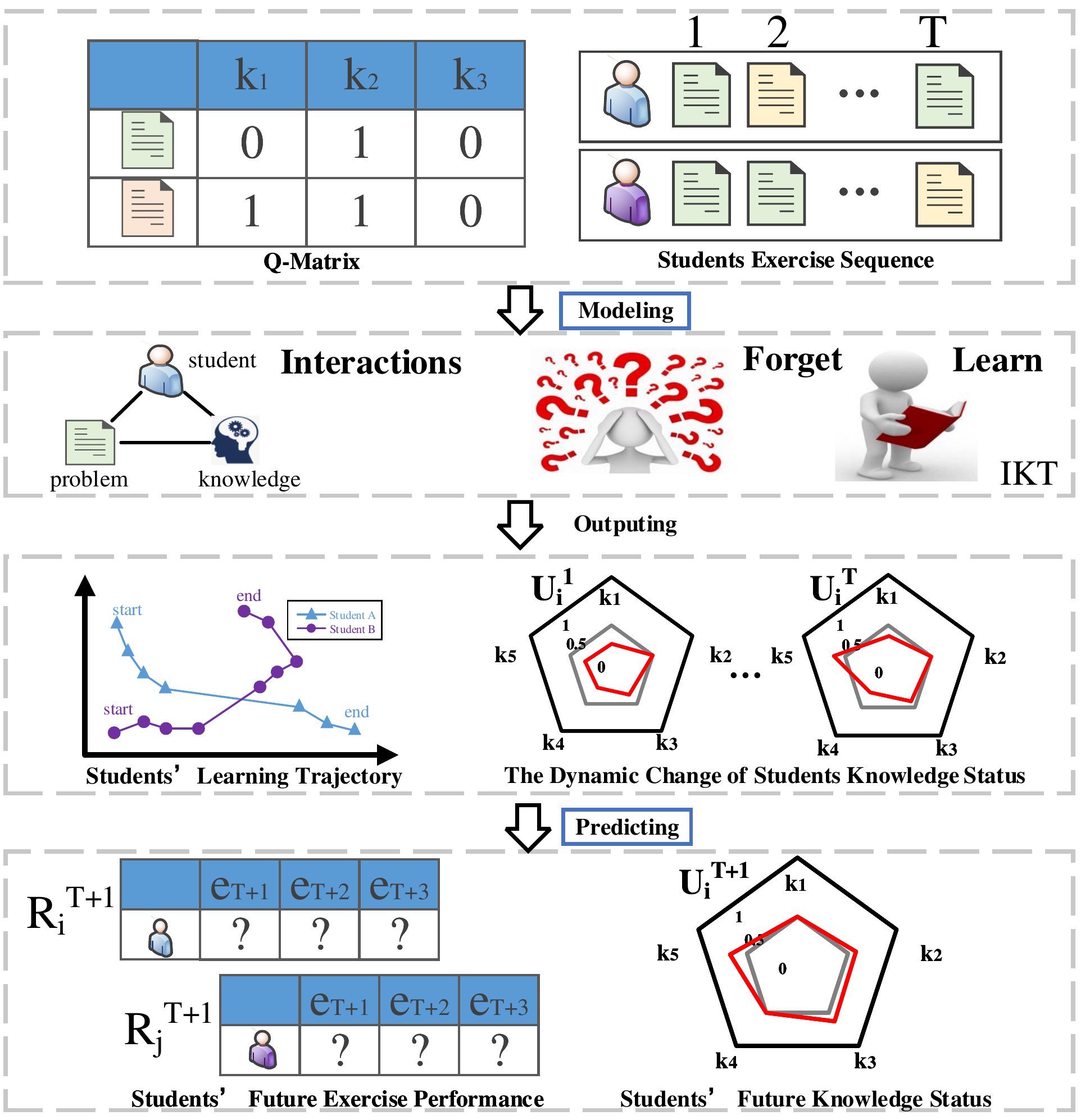}}
\caption{\label{fig:IKTFramework} The Framework of the TRACED model.}
\end{figure}

\subsection{Modeling Students' Exercise Responses}
\label{sec:orgb5a2624}
Students' performance is affected by many aspect, such as, student knowledge concept mastery, the knowledge concept contained in the problem and the similarity between difference knowledge concept. 
To modeling these different factors, inspired by factor analysis-based knowledge tracing models \cite{DBLP:conf/aaai/VieK19,DBLP:conf/edm/ChoffinPBV19}, 
we design three interactive strategies for modeling students' exercise responses, namely, 
strategies based on student problem interactions (UE), problem knowledge interactions (EK), and knowledge knowledge interactions (KK). UE, EK, and KK interactions are shown in Eq.\ref{eq:UEKK}. With these three interactive strategies, we predict students' exercise responses, and distributed representations of knowledge concepts and problems can be learned by exploring the three connections: a) connections between students and problems; b) connections between problems and concepts; and c) connections among knowledge concepts.
The proposed logarithmic linear model, which incorporates the three interactive strategies, is defined as follows: 
\begin{equation}
\label{eq:UEKK}
\begin{aligned}
&logit{(p(\hat{r}_{i,t} = 1 | u_i^t, e_{i,t}))} = \mu + w_{e, e_{i,t}} \\
& + \sum\limits_{k=1}^K Q_{e_{i,t}, k} w_{c, k}  + \underbrace{\sum\limits_{k_1=1}^K \sum\limits_{k_2=1}^K Q_{e_{i,t},k_1} Q_{e_{i,t},k_2} E_{c,k_1} E_{c,k_2}}_{\text{KK interactions}}\\
& + \underbrace{(\sum\limits_{k=1}^{K} u_{i,k}^t E_{c,k}) E_{e, e_{i,t}}}_{\text{UE interactions}} + \underbrace{\sum\limits_{k=1}^K Q_{e_{i,t},k} E_{e,e_{i,t}} E_{c,k}}_{\text{EK interactions}} \\
\end{aligned}
\end{equation}
where $p(\hat{r}_{i,t} = 1 | u_i^t, e_{i,t})$ represents the probability of student with knowledge concept mastery $u_i^t$ correctly answers the problem $e_{i,t}$ without considering slipping and guessing factors, \(logit (x) = \frac{x}{1-x}\), \(\hat{r}_{i,t}\) is student \(i\)'s exercise response in the \emph{t}-th exercise record; \(u_i^t\) is student \(i\)'s knowledge concept mastery at the time of the \emph{t}-th exercise record; \(\sum\limits_{k=1}^K u_{i,k}^t E_{c,k}\)  is the distributed representation of student \(i\) in the \emph{t}-th exercise record; \(E_{c,k}, E_{e,e'} \in \mathbb{R}^{d_e}\) are the distributed representations of knowledge concepts \(k\) and problems \(e'\); \(\mu\) is a global bias; \(w_e\), and \(w_{c}\) are the biases for the problems and knowledge concepts, respectively; and \(Q_{e,k}\) indicates whether problem \(e\) is related to knowledge concept \(k\).

\begin{figure}[!tbp]
\centering{
\includegraphics[width=0.48\textwidth]{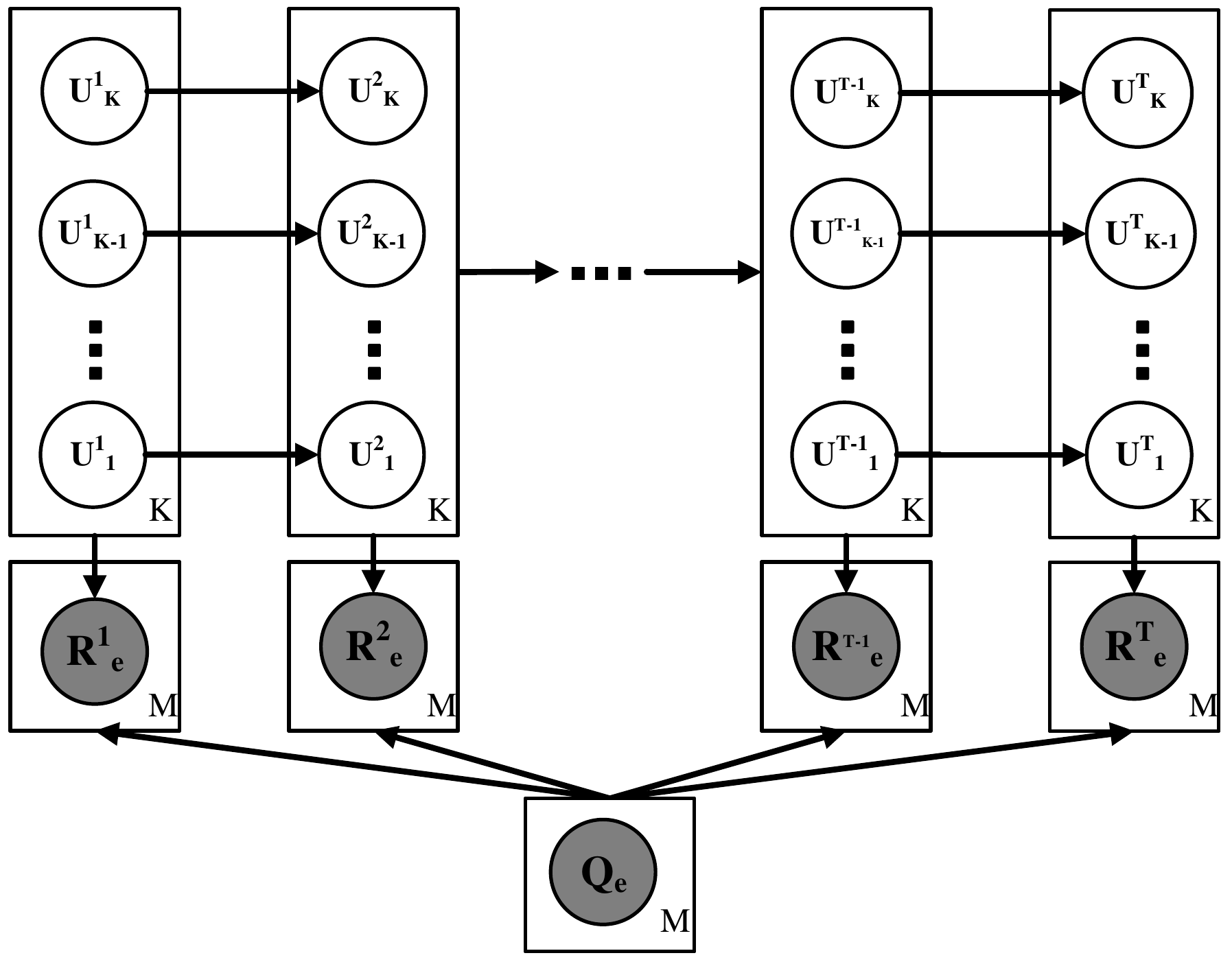}}
\caption{\label{fig:IKTFigure} Graphical representation of TRACED.}
\end{figure}

In addition, we consider the students' guessing and slipping behaviors when modeling the students' exercise responses as follows:
\begin{equation}
\begin{aligned}
&p \left( r_{i,t} = 1 | u_i^t \right) = p(\hat{r}_{i,t} = 1 | u_i^t) (1 - p(s_{e_{i,t}} = 1))\\
& \qquad \qquad \qquad+ p(\hat{r}_{i,t} = 0 | u_i^t) p(g_{e_{i,t}} = 1)\\
\end{aligned}
\end{equation}
The probability of making a mistake with problem \(e\) \(p(s_{e}=1)\) and the probability of correctly answering problem \(e\) by guessing \(p(g_{e}=1)\) are defined as:
\begin{equation}
\begin{aligned}
p \left( s_{e} = 1 \right) &= \frac{1}{1+e^{-\theta_{s,e}}}\\
p \left( g_{e} = 1 \right) &= \frac{1}{1+e^{-\theta_{g,e}}}\\
\end{aligned}
\end{equation}
where \(\theta_{s,e}\) and \(\theta_{g,e}\) are the guessing parameters and slipping parameters, respectively, for problem \(e\).

\subsection{Modeling Students' Knowledge Concept Mastery Probabilities Over Time}
\label{sec:orgb2040d8}
As students learn new knowledge concepts and forget the original knowledge concepts, the students' mastery of each knowledge concept varies over time. 
To better track students' knowledge concept mastery over time, we use a combination of learning and forgetting factors to model students' knowledge concept mastery. Student \(i\)'s mastery of knowledge concept \(k\) at the time $t$ is given as:
\begin{equation}
\begin{aligned}
p(u_{i,k}^t = 1) = &p(u_{i,k}^{t-1} = 1) (1 - p(F_k = 1)) \\
                   &\quad + p(u_{i,k}^{t-1} = 0) p(L_{k} = 1)\\
\end{aligned}
\end{equation}
where \(p(u_{i,k}^t = 1)\) is the probability that student \(i\) has mastered knowledge concept \(k\) at the time of the \emph{t}-th exercise; \(p(F_k=1)\) is the probability that a student will forget knowledge concept \(k\), and \(p(L_k=1)\) is the probability that a student will mastery knowledge concept \(k\) via learning. 
In this formula, the first component $p(u_{i,k}^{t-1} = 1) (1 - p(F_k = 1))$ represents the student has previously mastered knowledge concept $k$ and doesn't forget the knowledge concept $k$, and the second component $p(u_{i,k}^{t-1} = 0) p(L_{k} = 1)$ represents the student hasn't previously mastered knowledge concept $k$ but learned knowledge concept $k$. 

To facilitate the subsequent derivation, we define students' knowledge concept mastery in matrix form as follows:
\begin{equation}
\label{eq:pu}
p\left(u_{i, k}^t\right)=\left\{\begin{array}{lll}p\left(u_{i, k}^{t-1}\right) p\left(u_{i, k}^t \mid u_{i, k}^{t-1}\right) & \text { if } & t>1 \\ \pi(k) & \text { if } & t=1\end{array}\right .
\end{equation}
where \(p(u_{i,k}^t) = [p(u_{i,k}^t = 1), p(u_{i,k}^t = 0)]\) represents the probability of whether student $i$ masters knowledge concept $k$ at timestamp $t$, $\pi(k) = [\pi(k, 1), \pi(k, 0)]$ represents the probability of whether student masters knowledge concept $k$ when timestamp $t=1$, and \(p(u_{i,k}^t | u_{i,k}^{t-1})\) is the transfer matrix. 
Specifically, \(p(u_{i,k}^t | u_{i,k}^{t-1})\) is defined as: 
\begin{equation}
\label{eq:pu2u}
\begin{aligned}
&p(u_{i,k}^t | u_{i,k}^{t-1}) = 
\left [
\begin{array}{cc}
1 - p(F_k = 1) & p(F_k = 1) \\
p(L_{k} = 1)   & 1 - p(L_{k} = 1) \\
\end{array}
\right ]
\end{aligned} 
\end{equation}

Inspired by the learning curve and forgetting curve \cite{anzanello2011learning,von2007understanding}, we design the forgetting probability \(p(F_k = 1)\) and learning probability \(p(L_k = 1)\) for knowledge concept \(k\) based on the following principles: 
First, the longer the interval between two exercises on a particular knowledge concept, the higher the probability of forgetting that knowledge concept.
Second, the large the number of exercises for a particular knowledge concept within a short time, the higher the probability of learning that knowledge concept. 
\(p(F_k = 1)\) and \(p(L_k = 1)\) as follow:
\begin{align}
& p(F_k = 1) = \sigma (\frac{\Delta \tau_k}{\theta_{f,k}} + b_{f,k}) \label{eq:Fk} \\
& p(L_k = 1) = \sigma (\frac{\theta_{l1,k}*f_k}{f_k+\theta_{l2,k}} + b_{l,k}) \label{eq:Tk} 
\end{align}
where \(\sigma\) is sigmoid activation function; \(\Delta \tau_k\) is the time interval between the current exercise record for knowledge concept \(k\) and the previous record for knowledge concept \(k\), \(f_k\) denotes the exercise frequency for knowledge concept \(k\) within a specified time \(\Delta \hat{\tau}\); \(\theta_{f,k}\), \(\theta_{l1,k}\), and \(\theta_{l2,k}\) are the forgetting and learning parameters for knowledge concept \(k\);
\(b_{f,k}\) and \(b_{l,k}\) are the forgetting and learning biases for knowledge concept \(k\)
.

To better understand our proposed TRACED, we summarize the graphical representation of our model in Figure \ref{fig:IKTFigure}, where the shaded variables and unshaded variables indicate the observed variables and latent variables, respectively. Here, what we can observe are students' responses \(R\) with \(M\) problems and \(T\) exercises and the Q-matrix with \(K\) knowledge concepts.
The mastery \(U_{k}^{t}\) of knowledge concept \(k\) at the \emph{t}-th exercise depend on the mastery \(U_{k}^{t-1}\) of knowledge concept \(k\) at \emph{(t-1)}-th exercise, the time interval \(\Delta \tau_k\) between \(U_k^t\) and \(U_k^{t-1}\) and the exercise frequency \(f_{k}\) within \(\Delta \hat{\tau}\).
The response \(R_{e}^{t}\) about problem \(e\) at the \emph{t}-th exercise is influenced by problem information \(Q_{e}\), guessing factor \(g_{e}\), slipping factor \(s_{e}\) and knowledge concept mastery \(U^t\) at the \emph{t}-th exercise.
\subsection{Model Inference}
\label{sec:org18a4951}
In this section, we first infer the joint probability \(p(u_i, r_i)\) and marginal probability \(p(r_{i})\) and then explain the technical challenges encountered in model learning. Then, we detail the solutions proposed to address these technical challenges. 
\subsubsection{$p(u_i, r_i)$ and $p(r_i)$ Inference}
\label{sec:orgc78cd47}
We obtain the joint distribution of the observed and hidden variables by combining Eqs. \ref{eq:pu} and \ref{eq:pu2u}.
\begin{equation}
\label{eq:pur}
\begin{aligned}
&p(u_i, r_i) = p(u_i^1) \prod\limits_{t=2}^{T} p(u_i^t | u_i^{t-1}) \prod\limits_{ t=1 }^{ T } p(r_{i,t} | u_i^t) \\
&= \prod\limits_{k=1}^{K} p(u_{i,k}^1) \prod\limits_{t=2}^{T} \prod\limits_{k=1}^{K} p(u_{i,k}^t | u_{i,k}^{t-1}) \prod\limits_{ t=1 }^{T} p(r_{i,t} | u_{i,1}^t,...,u_{i,K}^t )
\end{aligned}
\end{equation}

By using maximum likelihood estimation to learn model parameters, we need to maximize the probability of observations (maximize the marginal probability distribution \(p(r_i)\)). The marginal probability distribution \(p(r_i)\) can be obtained as:
\begin{equation}
\label{eq:pr}
 \begin{aligned}
 p(r_i) &= \sum\limits_{u_i} p(u_i, r_i ) \\
 &= \sum\limits_{u_i} (p(u_{i}^1 ) \prod\limits_{ t=2 }^{T} p(u_i^t | u_i^{t-1}) \prod\limits_{ t=1 }^{ T } p(r_{i,t} | u_i^t))\\
 &= \sum\limits_{u_i} (p(u_{i}^1) p(r_{i,1} | u_i^1) \prod\limits_{ t=2 }^{ T } p(u_i^t | u_i^{t-1}) p(r_{i,t} | u_i^{t}))\\
 &=(\sum\limits_{u_i^{1}} p(u_i^1) p(r_{i,1}|u_i^1))*...* (\sum\limits_{u_i^{T}} p(u_i^T | u_i^{T-1}) p(r_{i,T}|u_i^T))\\
 \end{aligned}
\end{equation}
where \(\sum\limits_{u_i^{t}} p(u_i^t) p(r_{i,t}|u_i^t)\) is defined as follows:
\begin{equation}
\begin{aligned}
&\sum\limits_{u_i^{t}} p(u_i^t) p(r_{i,t}|u_i^t) = \sum\limits_{u_{i}^t} p(u_{i,1}^t) ... p(u_{i,K}^t) p(r_i | u_{i,1}^t,...,u_{i,K}^{t})
\end{aligned}
\end{equation}

Due to \emph{explaining away problem} (observation \(r_{i,t}\) depends on multiple hidden variables \(u_{i,1}^t, u_{i,2}^t ,..., u_{i,K}^{t}\) \cite{204911}, we cannot use the forward algorithm \cite{rabiner1989tutorial} to reduce the time complexity; hence, the time complexity of calculating \(P(r_i)\) is \(O(2^K*T)\), which is unacceptably high. Therefore, we cannot use maximum likelihood estimation for parameter optimization.
By using the Expectation-Maximization (EM) algorithm to learn model parameters, we need to maximize \(\sum\limits_{u} p(u | r) \log  p(u, r)\). The posterior distribution \(p(u_i | r_i)\) can be obtained by combining Eq.\ref{eq:pur} and Eq.\ref{eq:pr} as follows:
\begin{equation}
\label{eq:101}
\begin{aligned}
p (u_i| r_i) = \frac{p(u_i, r_i)}{p(r_i)}
\end{aligned}
\end{equation}
Since the time complexity of calculating the posterior distribution \(p(u_i | r_i)\) is also \(O(2^K * T)\),   
we cannot quickly sample the latent variables \(u_i\). Therefore, we cannot use the EM algorithm for parameter optimization.

To solve this problem, we improve an LSTM network to approximate the posterior distribution and propose a heuristic algorithm to learn the model parameters.

\subsubsection{Approximate Inference By Using LSTM}
\label{sec:orgc92de86}
We propose an LSTM-based network to approximate the posterior distribution \(p(u_i | r_i)\) as:
\begin{equation}
\label{eq:qphi}
\begin{aligned}
&q_{\phi}(u_{i}^{t}) = \sigma(W_q*LSTM(\tilde{x}_{i,t})+b_{q})\\
\end{aligned}
\end{equation}
where \(q_{\phi}(u_i^t) \in \mathbb{R}^{K}\) denotes the approximate result of the posterior distribution \([p(u_{i,1}^t|r_{i,t}),\) \(...,p(u_{i,K}^{t}|r_{i,t})]\),
\(\phi\) denotes all the training parameters in \(q_{\phi}\), and \(\sigma\) is the sigmoid activation function.

To better utilize and preserve the information of exercise responses $r_{i,t}$, we extend the students' exercise responses \(r_{i,t}\) to a feature vector \(\mathbf{0} = (0,0,...,0)\) with the same \(2d_e\) dimensions as the exercise embedding \(x_{i,t}\) and then learn the combined input vector \(\tilde{x}_{i,t} \in \mathbb{R}^{4d_e+K}\) as:
\begin{equation}
\begin{aligned}
&\widetilde{x}_{i,t}=\left\{\begin{array}{ll}
{\left[x_{i,t} \oplus \mathbf{0} \oplus p(u_{i}^t)\right]} & \text { if } r_{i,t}=1 \\
{\left[\mathbf{0} \oplus x_{i,t} \oplus p(u_{i}^t)\right]} & \text { if } r_{i,t}=0 \\
\end{array}\right. \\
\end{aligned}
\end{equation}
where \(\oplus\) is the operation of concatenating two vectors, and \(p(u_{i}^t) = [p(u_{i,1}^{t}),...,p(u_{i,K}^{t})]\) can be obtained from Eq.\ref{eq:pu}. By this way, LSTM can better identify and utilize exercise response information $r_{i,t}$ from embedding vector $\widetilde{x}_{i,t}$. And the exercise embedding \(x_{i,t}\) is defined as:
\begin{align}
& x_{i,t} = [E_{e,e_{i,t}}, \hat{E}_{e,e_{i,t}}] \label{eq:17_1}\\
& \hat{E}_{e,e_{i,t}} = \sum\limits_{j=1}^K Q_{e_{i,t}, j} E_{k,j} \label{eq:17_2}
\end{align}
where \(E_{k,j}, E_{e, e_{i,t}} \in \mathbb{R}^{d_e}\) are the distributed representations of the knowledge concept $j$ and problem $e_{i,t}$, and $Q_{e_{i,t},j}$ represents whether problem $e_{i,t}$ contains the knowledge concept $j$.

With the combined student \(i\)'s posterior information sequence \({\tilde{x}_{i,1},...,\tilde{x}_{i,T}}\), the hidden state \(h_{i,t} \in \mathbb{R}^{d_h}\) at t-th exercise is updated based on current combined input \(\tilde{x}_{i,t}\) and previous student state \(h_{i,t-1}\) as follows.
\begin{equation}
\begin{aligned}
i_{i,t} &=\sigma\left(\mathbf{Z}_{\mathbf{\tilde { x }i}} \tilde{x}_{i,t}+\mathbf{Z}_{\mathbf{h} \mathbf{i}} h_{i,t-1}+\mathbf{b}_{\mathbf{i}}\right) \\
f_{i,t} &=\sigma\left(\mathbf{Z}_{\widetilde{\mathbf{x}} \mathbf{f}} \tilde{x}_{i,t}+\mathbf{Z}_{\mathbf{h} \mathbf{f}} h_{i,t-1}+\mathbf{b}_{\mathbf{f}}\right) \\
o_{i,t} &=\sigma\left(\mathbf{Z}_{\tilde{\mathbf{x}} \mathbf{o}} \tilde{x}_{i,t}+\mathbf{Z}_{\mathbf{h} \mathbf{o}} h_{i,t-1}+\mathbf{b}_{\mathbf{o}}\right) \\
c_{i,t} &=f_{i,t} \cdot c_{i,t-1}+i_{i,t} \cdot \tanh \left(\mathbf{Z}_{\tilde{\mathbf{x}} \mathbf{c}} \tilde{x}_{i,t}+\mathbf{Z}_{\mathbf{h} \mathbf{c}} h_{i,t-1}+\mathbf{b}_{\mathbf{c}}\right) \\
h_{i,t} &=o_{i,t} \cdot \tanh \left(c_{i,t}\right)
\end{aligned}
\end{equation}
where \(\mathbf{Z_{\tilde{x}*}} \in \mathbb{R}^{d_h \times (4d_e+K)}\), \(\mathbf{Z}_{h*} \in \mathbb{R}^{d_h \times d_h}\) and \(\mathbf{b_{* }} \in \mathbb{R}^{d_h}\) are the parameters in LSTM.

\begin{algorithm}[htbp]
\caption{The training algorithm of TRACED}
\label{algorithm:KCRLESTraining}
\begin{algorithmic}[1]
\Require  
    Student exercise sequences $S = \left\{ e, t, r \right\}$.
\Ensure 
    The trained parameters $\phi$, and $\theta$.
\For {iteration = 1 to IterNum}
   \For {i = 1 to N}   
      \For {t = 1 to T}
          \State Sample ${u_{i}^{t}}' \sim q_{\phi}(u_i^t | r_{i,t})$;
      \EndFor
   \EndFor
   \State Update $\theta = \arg \min\limits_{\theta} - \frac{1}{N} \sum\limits_{i=1}^{N} \log p(u_i',r_{i})$;
   \For {i = 1 to N}
      \For {t = 1 to T}
          \State Sample ${u_i^t}' \sim p(u_i^t | {u_i^{t-1}}'))$;
          \State Sample ${r_{i,t}}' \sim p(r_{i,t}|{u_i^t}')$;
      \EndFor
   \EndFor
   \State Update $\phi = \arg \min\limits_{\phi} - \frac{1}{N} \sum\limits_{i=1}^N \log q_{\phi}({u_i}' | {r_i}')$;\
\EndFor
\end{algorithmic}
\end{algorithm}
\subsection{Model Learning}
\label{sec:orgf38d0ea}
This section explains in detail how the model's parameters are learned. The process of TRACED parameter learning is shown in Algorithm \ref{algorithm:KCRLESTraining}.

Inspired by \cite{hinton1995wake}, parameter learning is divided into the wake phase and the sleep phase. 
In the wake phase, we use \(q_{\phi}(u_i|r_i)\) to sample the hidden variables \(u_i\) and then optimize the parameters \(\theta\) of \(p(u_i, r_i)\) by using the \(u_i\) obtained via sampling.
In the sleep phase, we use \(p(u_i, r_i)\) to sample the hidden variables \(u_i\) and observed variables \(r_i\) and then optimize the parameters \(\phi\) of \(q_{\phi}(u_i | r_i)\) by the \(u_i\) and \(r_i\) obtained via sampling. In addition, Wake Phase and Sleep Phase both use real data, sample observe and hidden variables based on real data, and use real datas and sampled datas to train the model. 

\subsubsection{The loss function in the Wake Phase.}
\label{sec:org0ba821f}
In the Wake Phase, we utilize an LSTM-based network to approximate the posterior distribution \(p(u | r, \theta)\) and learn the parameters of TRACED by maximizing Evidence Lower BOund (ELBO). 
The derivation process of ELBO is as follows.
\begin{equation}
\begin{aligned} 
 &\log p(r | \theta) = \log \frac{p(u, r| \theta)}{p(u|r, \theta)}\\
 &= \sum \limits_u q_\phi (u | r) \log \frac{p(u, r| \theta)}{q_{\phi}(u | r)} \frac{q_{\phi}(u | r)}{p(u|r, \theta)}\\
 &= \underbrace{\sum\limits_{u} q_{\phi} (u | r) \log p(u|r, \theta) + H[q_{\phi}]}_{\text{Evidence Lower Bound (ELBO)}} \\
 & \qquad + KL (q_{\phi} (u | r) || p(u| r, \theta))\\
\end{aligned}
\end{equation}
where KL represents Kullback-Leibler divergence, and \(H[q_{\phi}] = - \sum\limits_u q_\phi(u | r) \log q_\phi(u | r)\). 
The derivation process of the loss function in the Wake Phase is expressed as follows:
\begin{equation}
\begin{aligned}
 \theta &= \arg \max_{\theta} ELBO\\
 &= \arg \max_{\theta} \sum\limits_{u} q_{\phi} (u | r) \log p(u|r, \theta) + H[q_{\phi}]\\
 &\propto \arg \max_{\theta} \sum\limits_{u} q_{\phi} (u | r) \log p \left( u, r| \theta \right)\\
 &= \arg \min_{\theta} - \frac{1}{N} \sum\limits_{i=1}^N \log p \left( u_i', r_i| \theta \right)
\end{aligned}
\end{equation}
where \(\theta\) represents all parameters of \(p(u,r|\theta)\), \(u_i'\) is sampled by \(q_{\phi}(u_i | r_i)\), and \(r_i\) is the real data.

\subsubsection{The loss function in the Sleep Phase.}
\label{sec:org5042490}
In the Sleep Phase, we learn the parameters \(\phi\) of the LSTM network. We minimize \(KL(p(u|r, \theta) || q_{\phi} (u|r, \phi))\) to better approximate the posterior distribution \(p(u|r, \theta)\). The derivation process of the loss function in the Sleep Phase is defined as follows.

\begin{equation}
\begin{aligned}
 \phi &= \arg \min_{\phi} KL(p( u | r, \theta) || q_{\phi} (u | r, \phi) )\\
 &\propto \arg \max_{\phi} \sum\limits_{u} p(r | \theta) p(u | r, \theta) \log \frac { q_{\phi} (u | r)}  {p(u | r, \theta)}\\
 & \quad- \sum\limits_{u} p(u,r| \theta) \log p(u | r, \theta) \\ 
 &\propto \arg \max_{\phi} E_{u,r \sim p(u, r | \theta)} (q_{\phi} (u | r))\\
 &= \arg \min_{\phi} -\frac{1}{N} \sum\limits_{i=1}^N \log q_{\phi}(u_i' | r_i')\\
\end{aligned}
\end{equation}
where \(\phi\) represents all parameters of \(q_{\phi}(u_i | r_i)\), and \(u_i'\) and \(r_i'\) are sampled by \(p(u_i, r_i | \theta)\).

\subsection{Predicting Student Future Knowledge Mastery Probabilities and Performance}
\label{sec:orgc51de35}
Before predicting students' future performance, we must predict students' future knowledge concept mastery probabilities. Given the previous \(t\) exercise records of student \(i\), we predict student \(i\)'s knowledge concepts mastery when the (t+1)-th exercise is performed at time \(\tau_{i,t+1}\) as:

\begin{equation}
\begin{aligned}
&p(u_i^{t+1} | r_{i,1}, r_{i,2}, ..., r_{i,t}, e_i, \tau_i)\\
&\approx \sum\limits_{u_i^t} q_{\phi} (u_i^t | r_{i,1}, ..., r_{i,t}, e_i, \tau_i) p(u_i^{t+1} | u_i^t, \tau_i)\\
&= (\sum\limits_{u_{i,1}^{t}} q_{\phi} (u_{i,j}^t | r_{i,1}, ..., r_{i,t}, e_i, \tau_i) p(u_{i,1}^{t+1} | u_{i,1}^t, \tau_i))* ... *\\
& \quad (\sum\limits_{u_{i,K}^{t}} q_{\phi} (u_{i,K}^t | r_{i,1}, ..., r_{i,t}, e_i, \tau_i) p(u_{i,K}^{t+1} | u_{i,K}^t, \tau_i))
\end{aligned}
\end{equation}
where student \(i\)'s exercise time \(\tau_i = \left\{ \tau_{i,1},...,\tau_{i,t+1}\right\}\) and student \(i\)'s exercise problem \(e_i = \left\{ e_{i,1}, ..., e_{i,t} \right\}\).

Furthermore, we can predict the probability that student \(i\) correctly answers problem \(e_{i,t+1}\) at time \(\tau_{i,t+1}\) as:
\begin{equation}
\begin{aligned}
&p(r_{i,T+1}| r_{i,1}, r_{i,2}, ... , r_{i,T}, e_{i}, \tau_i)\\
&= \frac{p(r_{i,T+1}, u_i^{T+1}| r_{i,1}, r_{i,2}, ... , r_{i,T}, e_i, \tau_i)}{p(u_i^{T+1}| r_{i,1}, r_{i,2}, ..., r_{i,T}, r_{i, T+1}, e_i, \tau_i)}\\ 
&= \frac{p(u_i^{T+1}| r_{i,1}, r_{i,2}, ... , r_{i,T}, e_i, \tau_i) p(r_{i, T+1} | u_i^{T+1}, \tau_i)}{p(u_i^{T+1}| r_{i,1}, r_{i,2}, ...,r_{i,T}, r_{i,T+1}, \tau_i)} \\
&\approx \frac{ p(r_{i,T+1} | u_i^{T+1}, e_i) (\sum\limits_{u_i^T} q_{\phi} (u_i^T | r_{i,1\sim T}, e_i, \tau_i) p(u_i^{T+1} | u_i^T, \tau_i))}{q_{\phi} (u_i^{T+1} | r_{i,1 \sim T+1}, e_i, \tau_i)} 
\end{aligned}
\end{equation} 
Since the time complexity of \(\sum\limits_{u_i^T} q_{\phi} (u_i^T | r_{i,1\sim T}, e_i, \tau_i)\) \(p(u_i^{T+1} | u_i^T, \tau_i)\) is \(O(2^K)\), we propose to improve an LSTM network to approximate \(p(r_{i,T+1}| r_{i,1}, r_{i,2}, ... \) \(, r_{i,T}, e_{i}, \tau_i)\) by the distributed representations of students' exercise records. Finally, we predict students' future exercise responses as follows.
\begin{align}
& y_p(i,t+1) = \sigma(W_n*LSTM(\hat{x}_{i,t+1})+b_{n})\\
& \hat{x}_{i,t+1}= \left[x_{i,t+1} \oplus q_{\phi}(u_{i}^{t}) \oplus p(u_{i}^{t+1})\right] 
\end{align}
where \(\hat{x}_{i,t+1}\) represents the distributed representation of student \emph{i}'s \emph{t+1}-th exercise record without exercise response, which contains all the information needed to calculate \(p(r_{i,T+1}| r_{i,1}, r_{i,2}, ... , r_{i,T}, e_{i}, \tau_i)\) (priori, posterior, and distributed representation); \(x_{i,t+1}\), \(q_{\phi}(u_i^{t})\), and \(p(u_i^{t+1})\) can be obtained via Eqs. \ref{eq:17_1}, \ref{eq:17_2}, \ref{eq:qphi}, and \ref{eq:pu}, respectively; and \(y_{p}(i,t+1)\) represents the probability that student \(i\) will answer correctly during the \emph{t+1}-th exercise.

\subsection{Predicting Relationships between Concepts}
\label{sec:org1ea799b}
Given part of a knowledge concept graph, we can build a supervised model to fill in the gaps in the graph. In this paper, this goal is simplified to the prediction of only the inclusion relationships between knowledge concepts.
The inclusion relationship between knowledge concepts \(i\) and \(j\) is predicted as follows:
\begin{equation}
\label{equ:PredictingRelationshipsbetweenKnowledgeConcepts}
\begin{aligned}
&h_r = tanh(W_{r,h} * E_r + b_{r,h})\\
&y_{r} (i,j) = \sigma(W_{r,o} * h_r + b_{r,o})
\end{aligned}
\end{equation}
where \(E_r = [E_{e,i}, E_{e,j}, E_{e,i} -  E_{e,j}]\) and \(W_{r,h}\), \(W_{r,o}\), \(b_{r,h}\), and \(b_{r,o}\) are the training parameters of the fully connected neural network.

\subsection{Predicting Concepts contained by Problems}
\label{sec:orga2021d1}
Given part of a Q-matrix, we can use the given Q-matrix to train a supervised model to judge whether a certain problem contains a certain knowledge concept. This model can be used to fill in the missing values of the Q-matrix and to fuzzify the Q-matrix by replacing the original 01 matrices with a continuous probability matrix. 
We predict the inclusion relationships between knowledge concept and problem as follows:
\begin{equation}
\label{eq:PredictingConceptsInProblems}
\begin{aligned}
&h_Q = tanh(W_{Q,h}*E_Q + b_{Q,h})\\
&y_{Q} (i,j) = \sigma(W_{Q,o} * h_Q + b_{Q,o})
\end{aligned}
\end{equation}
where \(E_Q = [E_{e,i}, E_{k,j}, E_{e,i} - E_{k,j}]\) and \(W_{Q,h}\), \(W_{Q,o}\), \(b_{Q,h}\), and \(b_{Q,o}\) are the training parameters of the fully connected neural network.

%% file: src/5-Experiment.tex
\section{Experiment}
\label{sec:orgf5eee29}
\subsection{Datasets}
\label{sec:org033968f}
We employ four real-world datasets: POJ, HDU, algebra06 and algebra08. The HDU and POJ datasets were obtained from the Hangzhou Dianzi University Online Judge platform (\url{http://acm.hdu.edu.cn}, accessed on October 22 2021) and the Peking University Online Judge platform (\url{http://poj.org}, accessed on 2 January 2023). Moreover, the algebra06 and algebra08 datasets obtained from the KDD Cup 2010 EDM Challenge (\url{https://pslcdatashop.web.cmu.edu/KDDCup/downloads.jsp}, accessed on 2 January 2023).
\begin{table}[!tbp]
\centering{
\caption{Statistics of the Datasets\label{table:statisticsofdatasets} }
\begin{tabular}{c|c|c|c|c}
\hline
 Dataset     &  HDU      &    POJ        &   algebra06      &   algebra08       \\
\hline
 \#Student   & 9,859     &    3,507      &    1,072         &     2,385         \\
 \#Problem   & 2,101     &     970       &    1,218         &      736          \\
 \#Records   & 1,042,661 &   288,594     &   1,199,438      &    1,895,107      \\
 \#Concepts  &  193      &    146        &     319          &      304          \\
 Avg.rec     &  105      &     82        &    1,118         &      795          \\
\hline
\end{tabular}}
\end{table}
The HDU and POJ dataset includes records submitted from June 2018 to November 2018, the algebra06 dataset includes records submitted from October 2006 to February 2007, and the algebra08 dataset includes records submitted from September 2008 to January 2009. 
Students are allowed to resubmit their codes for a problem until they pass. 
We filter out students with fewer than 30 records and an acceptance rate of less than 10\%, as well as problems with fewer than 30 records. 
After filtering, the statistics of the datasets are shown in Table \ref{table:statisticsofdatasets}, and Avg.rec in Table \ref{table:statisticsofdatasets} represents the average number of students exercise records. 

Specifically, numerous problems in the HDU and POJ datasets lacked knowledge concept labels. The knowledge concept of the problem in the HDU and POJ dataset is highly consistent with the knowledge concept of the problem in ICPC (\url{https://icpc.global}, accessed on 2 January 2023). It covers a wide range. The granularity of knowledge concepts is coarse, such as dynamic programming, graph theory, and soon on. Therefore, we gathered five experts who have won the medal of the ICPC Asian regional contest to mark the problems with knowledge concept labels. If more than three experts mark a knowledge concept label for a certain problem, we add the label to the problem. 

\subsection{Evaluation Metrics}
\label{sec:orgcc7a275}
We evaluate the models for predicting students' future performance from regression and classification perspectives \cite{fogarty2005case,DBLP:conf/aaai/SuLLHYCDWH18}. 

For regression, we use the \emph{Mean Absolute Error} (MAE) and the \emph{Root-Mean-Square Error} (RMSE) to quantify the distance between the predicted scores and the actual ones. 

For classification, we select the \emph{Area Under the Receiver Operating Characteristic Curve} (AUC), \emph{Prediction Accuracy} (ACC), \emph{Precision} (PRE) and \emph{Recall} (REC) as measures of performance. The larger the values of these metrics are, the better the results are. The threshold adopted by the classification task in this paper is 0.5. 

Liu et al. \cite{liu2022learning} noted that various learning rate schedules can have a notable effect on performance. To ensure the fairness of the experiment, constant learning rate was utilized for all models. 
And all models are implemented by Python, and all experiments are run on a Linux server with eight 3.6 GHz Intel W-2123 CPUs, 64 G memory and an NVIDIA RTX-2080Ti GPU. 

\begin{table*}[!htbp]
  \small
  \caption{Results for Predicting Future Student Performance on the HDU and POJ datasets}
  \label{table:FuturePerformanceResultHDU}
  \centering
  \begin{tabular}{c|cccccc|cccccc}
  \hline
   & \multicolumn{6}{c|}{HDU} & \multicolumn{6}{c}{POJ}\\
  \hline
   Model       &  AUC & ACC & PRE & REC & RMSE & MAE & AUC & ACC & PRE & REC & RMSE & MAE\\
  \hline
  IRT   & 0.6329 & 0.6407 & 0.5652 & 0.3007 & 0.4741 & 0.4398  &  0.6067 & 0.6594 & 0.5642 & 0.1294 & 0.2206 & 0.4303  \\
  MIRT  & 0.6376 & 0.6410 & 0.5596 & 0.3285 & 0.4731 & 0.4493  &  0.6099 & 0.6602 & 0.5593 & 0.1486 & 0.2193 & 0.4403 \\
  AFM   & 0.5669 & 0.6155 & 0.5276 & 0.0426 & 0.4840 & 0.4669  &  0.5154 & 0.6488 & 0.3269 & 0.0108 & 0.2275 & 0.4546  \\
  PFA   & 0.6394 & 0.6349 & 0.6169 & 0.1417 & 0.4738 & 0.4488  &  0.5337 & 0.6506 & 0.5536 & 0.0215 & 0.2262 & 0.4523  \\
  KTM   & 0.6760 & 0.6619 & 0.6104 & 0.3423 & 0.4639 & 0.4291  &  0.6149 & 0.6603 & 0.5525 & 0.1683 & 0.2194 & 0.4340  \\
  DASH  & 0.6808 & 0.6644 & 0.6068 & 0.3705 & 0.4621 & 0.4223  &  0.6149 & 0.6603 & 0.5525 & 0.1683 & 0.2194 & 0.4340  \\
  DAS3H & 0.6794 & 0.6633 & 0.5957 & 0.3966 & 0.4627 & 0.4236  &  0.6084 & 0.6528 & 0.5148 & 0.1815 & 0.2210 & 0.4409  \\
  DKT   & 0.6986 & 0.6752 & 0.6224 & 0.4327 & 0.4136 & 0.4581  &  0.6601 & 0.6757 & 0.5627 & 0.2762 & 0.2012 & 0.4123 \\
  DKVMN & 0.6959 & 0.6761 & 0.6304 & 0.4126 & 0.4134 & 0.4589  &  0.6578 & 0.6804 & 0.5814 & 0.2642 & 0.2094 & 0.4121 \\
  AKT   & 0.7019 & 0.6805 & 0.6201 & 0.3715 & 0.4136 & 0.4544  &  0.5913 & 0.6618 & 0.5627 & 0.0894 & 0.2213 & 0.4392 \\
  TRACED   & \textbf{0.7328} & \textbf{0.7096} & \textbf{0.6412}  & \textbf{0.4346} & \textbf{0.4074} & \textbf{0.4489} & \textbf{0.6674} & \textbf{0.6962} & \textbf{0.5884} & \textbf{0.2846} & \textbf{0.2011} & \textbf{0.4094}\\
  \hline
  \end{tabular}
 \end{table*}

\subsection{Baselines for Comparison}
\label{sec:org47282c8}
We compare TRACED with the following eleven state-of-the-art methods with well-tuned parameters.  
\begin{itemize}
\item \emph{IRT} and \emph{MIRT} \cite{embretson2013item}: a popular cognitive diagnostic model, which discover students' knowledge levels through ranking with a logistic-like function. When d $>$ 0, IRT becomes MIRT, a variant of MIRT that considers a user bias.
\item \emph{AFM} and \emph{PFA} \cite{pavlik2009performance,cen2006learning}: the factor analysis models, which take account the number of attempts (AFM) and the number of positive and negative attempts (PFA). 
\item \emph{KTM} \cite{DBLP:conf/aaai/VieK19}: a factorization machine model, which encompasses IRT, AFM, and PFA as special cases. 
\item \emph{DASH} \cite{lindsey2014improving,mozer2016predicting}: a knowledge tracing model that bridges the gap between factor analysis and memory models. It stands for difficulty, ability, and student history.  
\item \emph{DAS3H} \cite{DBLP:conf/edm/ChoffinPBV19}: a framework based on factor analysis, which takes both memory decay and multiple knowledge concept tagging into account. 
\item \emph{DKT} \cite{DBLP:conf/nips/PiechBHGSGS15}: the first model to apply deep learning algorithms for knowledge tracing, which uses LSTM to track students’ knowledge mastery over time. 
\item \emph{DKVMN} \cite{DBLP:conf/www/ZhangSKY17}: a key-value memory networks, which exploit the relationship among the underlying knowledge and directly outputs a student’s knowledge proficiency. 
\item \emph{AKT} \cite{DBLP:conf/kdd/0001HL20}: a monotonic attention mechanism model, which predict students' future performance by relating a students' future responses to assessment problems to their past responses. 
\end{itemize}
We do not perform a comparison with FuzzyCDF \cite{DBLP:conf/ijcai/WuLLCSCH15}, DINA \cite{de2009dina}, and KPT \cite{DBLP:conf/cikm/ChenLHWCWSH17}. 
Because we regard exercise records of a student as complete data, which either appear in the training set or the test set, however, FuzzyCDF and DINA contain student parameters that need to be trained. Thus, a student’s exercise record must appear in both the training and test sets. The KPT model cannot predict student performance in real time. 
\subsection{Predicting Future Student Performance}
\label{sec:org15c97a2}
In this section, we compare the predictions of our model and other baseline models in students' future performance. The experimental parameters of TRACED are \(\Delta \hat{\tau} = 86400s\),  \(d_e = 20\), \(d_{h} = 2 * K\), \(d_{p} = 80 + 4 * K\) and the number of exercise record embedding dimensions is \(40 + 2 * K\). Besides, we randomly initialize all parameters in the TRACED to the Glorot normal \cite{pmlr-v9-glorot10a}.
To prevent overfitting, we add L2 regularization for TRACED. We perform 5-fold cross-validation for each dataset and average the results over five folds. For each fold, we use 80\% of the data for training and use the remaining 20\% for testing. 

The experimental results are shown in Table \ref{table:FuturePerformanceResultHDU} and \ref{table:FuturePerformanceResultKDD}.
It can be seen that the results of all models in the algebra06 and algebra08 datasets are much better than the results with the HDU and POJ datasets, because the average number of students' exercise submissions with the algebra06 and algebra08 datasets are much higher than that with the POJ and HDU datasets. Besides, learners in POJ and HDU datasets mostly learn to program independently, and they often submit repeated submissions to test the accuracy and efficiency of the code. Learners in algebra06 and algebra08 datasets are mostly after-school math exercises. Although learners will submit many times, the number of repeated submissions is less. 
Most models in the algebra06 and HDU datasets are much better than those in the algebra08 and POJ datasets. Because most models perform better in large datasets than in small datasets when using the same type of data set. 

The performance of TRACED is significantly better than that of other models.
The reason why the performance of TRACED is better than that of FAKT (DAS3H, KTM, etc.) is that the TRACED model can be regarded as a kind of DLKT for predicting students' future performance tasks.
DLKT is significantly better than FAKT for predicting students' future performance tasks.
The TRACED model is better than other DLKT models because TRACED does not directly use the original exercise records but uses the distributed representations of records to predict students' future performance. Compared with the original records, the distributed representations of exercise records contain a substantial amount of contextual information. 

\begin{table*}[htbp]
 \small
 \caption{Results for Predicting Future Student Performance on the algebra06 and algebra08 datasets}
 \label{table:FuturePerformanceResultKDD}
 \centering
 \begin{tabular}{c|cccccc|cccccc}
 \hline
  & \multicolumn{6}{c|}{algebra06} & \multicolumn{6}{c}{algebra08}\\
 \hline
  Model       &  AUC & ACC & PRE & REC  & RMSE & MAE &  AUC & ACC & PRE & REC & RMSE & MAE\\
 \hline 
 IRT   & 0.6663 & 0.8451 & 0.8477 & 0.9957 & 0.1244 & 0.2397 & 0.6668 & 0.8123 & 0.8148 & 0.9948 & 0.3798 & 0.2904  \\
 MIRT  & 0.6625 & 0.8455 & 0.8467 & \textbf{0.9979} & 0.1247 & 0.2577 & 0.6656 & 0.8123 & 0.8144 & \textbf{0.9956} & 0.3802 & 0.2998  \\
 AFM   & 0.6663 & 0.8451 & 0.8477 & 0.9957 & 0.1244 & 0.2597 & 0.6737 & 0.8288 & 0.8190 & 0.9862 & 0.3820 & 0.2927  \\
 PFA   & 0.7120 & 0.8418 & 0.8567 & 0.9761 & 0.1220 & 0.2319 & 0.7040 & 0.8143 & 0.8179 & 0.9918 & 0.3746 & 0.2806  \\
 KTM   & 0.7440 & 0.8484 & 0.8546 & 0.9890 & 0.1155 & 0.2298 & 0.7173 & 0.8161 & 0.8214 & 0.9883 & 0.3717 & 0.2762  \\
 DASH  & 0.7464 & 0.8512 & 0.8548 & 0.9927 & 0.1143 & 0.2425 & 0.7090 & 0.8142 & 0.8172 & 0.9930 & 0.3742 & 0.2934  \\
 DAS3H & 0.7328 & 0.8419 & 0.8580 & 0.9743 & 0.1227 & 0.2790 & 0.7234 & 0.8164 & 0.8214 & 0.9887 & 0.3704 & 0.2738 \\
 DKT   & 0.7513 & 0.8536 & 0.8497 & 0.9826 & 0.1124 & 0.2310 & 0.7462 & 0.8182 & 0.8315 & 0.9728 & 0.2638 & 0.3663 \\
 DKVMN & 0.7564 & 0.8579 & 0.8592 & 0.9910 & 0.1117 & 0.2284 & 0.7453 & 0.8188 & 0.8288 & 0.9785 & 0.2662 & 0.3662 \\
 AKT   & 0.7573 & 0.8621 & 0.8588 & 0.9954 & 0.1106 & 0.2193 & 0.7173 & 0.8090 & 0.8158 & 0.9857 & 0.2750 & 0.3769 \\
 TRACED   & \textbf{0.7604} & \textbf{0.8623} & \textbf{0.8596} & 0.9957 & \textbf{0.1098} & \textbf{0.2154} & \textbf{0.7724} & \textbf{0.8336} & \textbf{0.8496} & 0.9894 & \textbf{0.2539} & \textbf{0.3659}\\
 \hline
 \end{tabular}
\end{table*} 
The performance of FAKT models is constantly improving with an increase in the time features extracted by humans. The performance is reflected in the performance comparison of DAS3H, DASH, KTM, PFA, AFM MIRT and IRT (DAS3H $>$ DASH $>$ KTM $>$ PFA, AFM $>$ MIRT $>$ IRT).
Specifically, AFM and PFA are the factor analysis models considering the number of attempts and positive and negative attempts. Since the number of attempts in the HDU and POJ datasets cannot accurately reflect whether learners master knowledge concepts, the performance of AFM and PFA in the HDU and POJ datasets is poor, even worse than IRT and MIRT. However, the performance of AFM and PFA in the algebra06 and algebra08 datasets is better than IRT and MIRT.
The performance gap among the three types of DLKT is tiny.
The reason why the performance of DKVMN is not much different from that of DKT may be that we may not need too many storage units to obtain better performance in these four datasets.
The reason for the unsatisfactory effect of AKT is that the AKT model is a model designed for problems with a single knowledge concept, and problems in the HDU, POJ, algebra06 and algebra08 datasets contain a substantial amount of knowledge concepts.
Besides, the reason for the poor performance of AKT in the POJ dataset is that the number of learners and submissions in the POJ dataset is too small. 

\subsection{Predicting Relationships Between Concepts}
\label{sec:org9798341}
Since we do not have experts in the field of algebra, we have not constructed a knowledge graph in the field of algebra. In the end, we chose to conduct the experiment of predicting the relationship between knowledge concepts in the HDU and POJ datasets. We use a knowledge concept graph to train a fully connected neural network to identify the inclusion relationships between knowledge concepts. The fully connected neural network has 30 hidden units, and the number of dimensions of the knowledge concept embeddings is 40; the network structure is shown in Eq. \ref{equ:PredictingRelationshipsbetweenKnowledgeConcepts}.

The sparsity of the knowledge concept graph leads to an extreme imbalance between the number of positive and negative samples in the data.  
To overcome this problem, we sample 2 negative samples for each positive sample, following the negative sampling approach proposed in \cite{mikolov2013distributed}. 
Finally, each data point is represented as \((k_1, k_2, r)\), where \(r\) represents the relationship between knowledge concepts \(k_1\) and \(k_2\) (\(r=1\) means that knowledge concept \(k_1\) includes knowledge concept \(k_2\); otherwise, \(r=0\)). 
For each fold, we use 80\% of the data as training data and the remaining 20\% as test data. The experimental results are shown in Table \ref{table:RelationshipResult}. 
The performance of the fully connected neural networks that consider the distributed representations of the knowledge concepts is much better than that of the networks that do not consider the distributed representations of the knowledge concepts. 
Moreover, the distributed representations based on both dynamic and static interactions yield better performance than the other methods for predicting the relationships between knowledge concepts. 

\begin{table}[!tbp]
  \small
  \caption{Results of Predicting Relationships Between Concepts}
  \label{table:RelationshipResult}
  \centering
  \begin{tabular}{m{3.1cm}  m{0.7cm}  m{0.7cm}  m{0.7cm}  m{0.7cm}}
  \hline
  Model       &  AUC & ACC & RMSE & MAE \\
  \hline 
  \multicolumn{5}{c}{HDU}\\
  \hline
  NN               & 0.780 & 0.683 & 0.390 & 0.282 \\  
  NN + EK,KK       & 0.829 & 0.715 & 0.371 & 0.302 \\ 
  NN + EK,UE       & 0.807 & 0.656 & 0.371 & 0.305 \\ 
  NN + KK,UE       & 0.812 & 0.698 & 0.385 & 0.308 \\ 
  NN + EK,KK,UE    & \textbf{0.848} & \textbf{0.746} & \textbf{0.360} & \textbf{0.300} \\ 
  \hline
  \multicolumn{5}{c}{POJ}\\
  \hline
  NN               & 0.713 & 0.619 & 0.463 & 0.353 \\  
  NN + EK,KK       & 0.751 & 0.609 & 0.412 & 0.352 \\ 
  NN + EK,UE       & 0.732 & 0.566 & 0.432 & 0.373 \\ 
  NN + KK,UE       & 0.751 & 0.604 & 0.413 & 0.354 \\ 
  NN + EK,KK,UE    & \textbf{0.812} & \textbf{0.768} & \textbf{0.393} & \textbf{0.349} \\ 
  \hline
  \end{tabular}
 \end{table}

 \begin{table}[!tbp]
  \small
  \caption{Results of Predicting Concepts of Problems}
  \label{table:PredictConceptResult}
  \centering
  \begin{tabular}{m{3.1cm}  m{0.7cm}  m{0.7cm}  m{0.7cm}  m{0.7cm}}
  \hline
  Model       & AUC & ACC & RMSE & MAE \\
  \hline 
  \multicolumn{5}{c}{HDU}\\
  \hline
  NN               & 0.688 & 0.530 & 0.458 & 0.377 \\  
  NN + EK,KK       & 0.759 & 0.683 & 0.417 & 0.363 \\ 
  NN + EK,UE       & 0.756 & 0.682 & 0.418 & 0.363 \\ 
  NN + KK,UE       & 0.753 & 0.668 & 0.422 & 0.378 \\ 
  NN + EK,KK,UE    & \textbf{0.764} & \textbf{0.670} & \textbf{0.416} & \textbf{0.365} \\ 
  \hline
  \multicolumn{5}{c}{POJ}\\
  \hline
  NN               & 0.678 & 0.511 & 0.463 & 0.379 \\  
  NN + EK,KK       & 0.769 & 0.706 & 0.416 & 0.375 \\ 
  NN + EK,UE       & 0.767 & 0.684 & 0.418 & 0.379 \\ 
  NN + KK,UE       & 0.763 & 0.707 & 0.427 & 0.399 \\ 
  NN + EK,KK,UE    & \textbf{0.772} & \textbf{0.717} & \textbf{0.413} & \textbf{0.366} \\ 
  \hline
  \end{tabular}
 \end{table}
 
The distributed representations that consider KK interactions show better performance for this task than those that consider EK interactions. 
In summary, using UE, EK, and KK interaction strategies can learn the relationship information between knowledge concepts, and the KK interaction strategy plays a more important role than the EK, UE interaction strategy in the task of learning knowledge concept relationship information.
The experimental results show that the distributed representation of knowledge concepts learned by TRACED contains effective information about the relationship between knowledge concepts.

\subsection{Predicting Concepts Contained by Problems}
\label{sec:org7a625c6}
We use the Q-matrix to train a fully connected neural network to judge whether a certain problem contains a certain knowledge concept. 
The fully connected neural network has 30 hidden units, and the number of dimensions of the knowledge concept embeddings is 40; the network structure is shown in Eq. \ref{eq:PredictingConceptsInProblems}. 
Due to the sparsity of the Q-matrix, we sample 2 negative samples for each positive sample. Finally, each data point is represented as \((e, k, r)\), where \(r = 1\) indicates that problem \(e\) is related to knowledge concept \(k\). 

For each fold, we use 80\% of the data for training and the remaining 20\% for testing. 
The experimental results are shown in Table \ref{table:PredictConceptResult}. 
Again, the performance of the fully connected neural networks that consider the distributed representations of the knowledge concepts is much better than that of the networks that do not consider the distributed representations of the knowledge concepts.
The distributed representations based on both dynamic and static interactions yield better performance than the other methods for predicting the knowledge concepts contained by problems.
Besides, the distributed representations that consider EK interactions show better performance for this task than those that consider KK interactions.
In summary, using UE, EK, and KK interaction strategies can learn the relationship information between knowledge concept and problem, and the EK interaction strategy plays a more important role than the KK, UE interaction strategy in the task of learning knowledge concept relationship information.
The experimental results show that the distributed representation of knowledge concepts learned by TRACED contains effective information about the relationship between knowledge concept and problem.

\begin{figure}[!ht]
  \centering
  \subfigure[the loss in wake phase]{\includegraphics[width=0.24\textwidth]{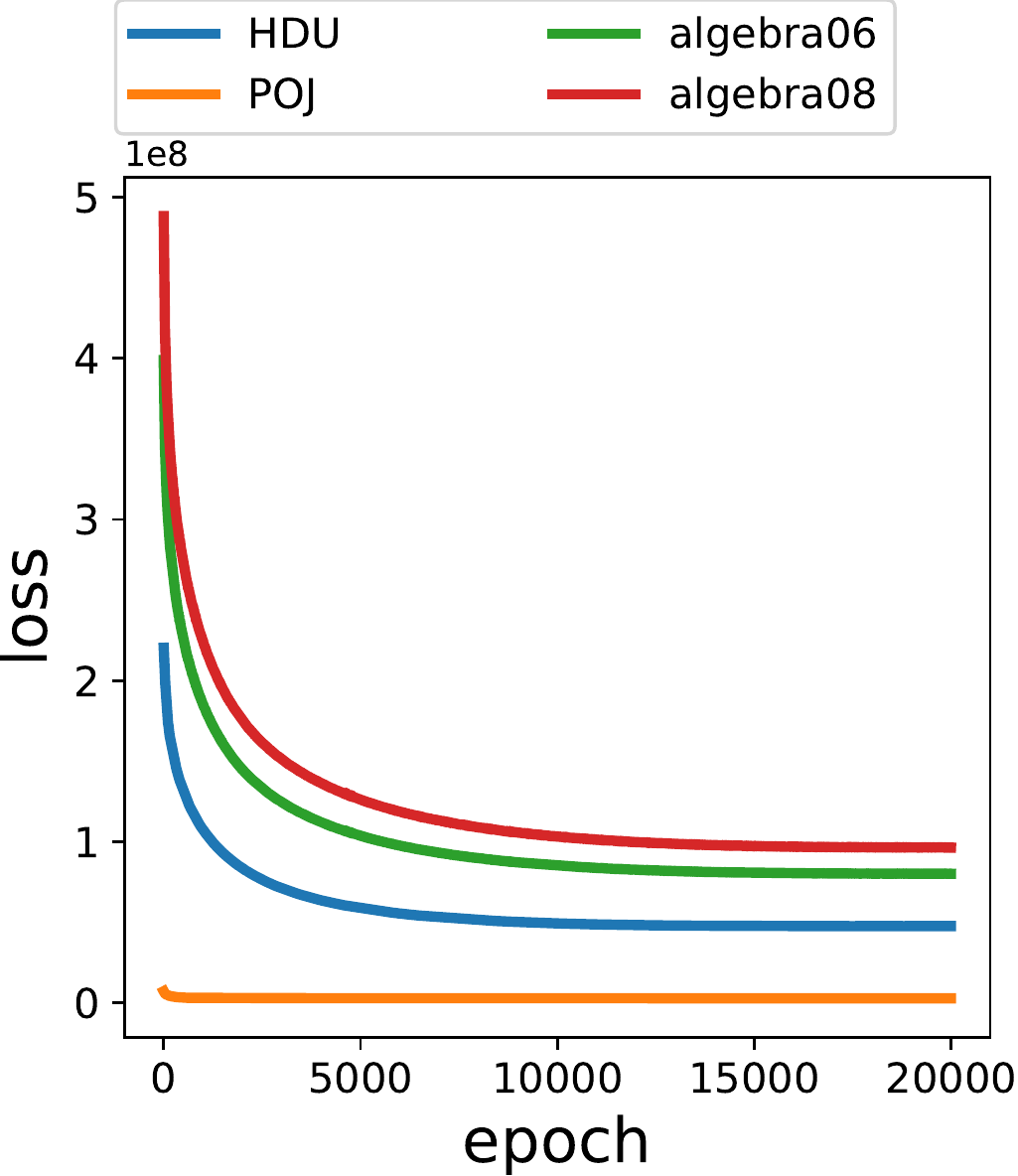}}
  \subfigure[the loss in sleep phase]{\includegraphics[width=0.24\textwidth]{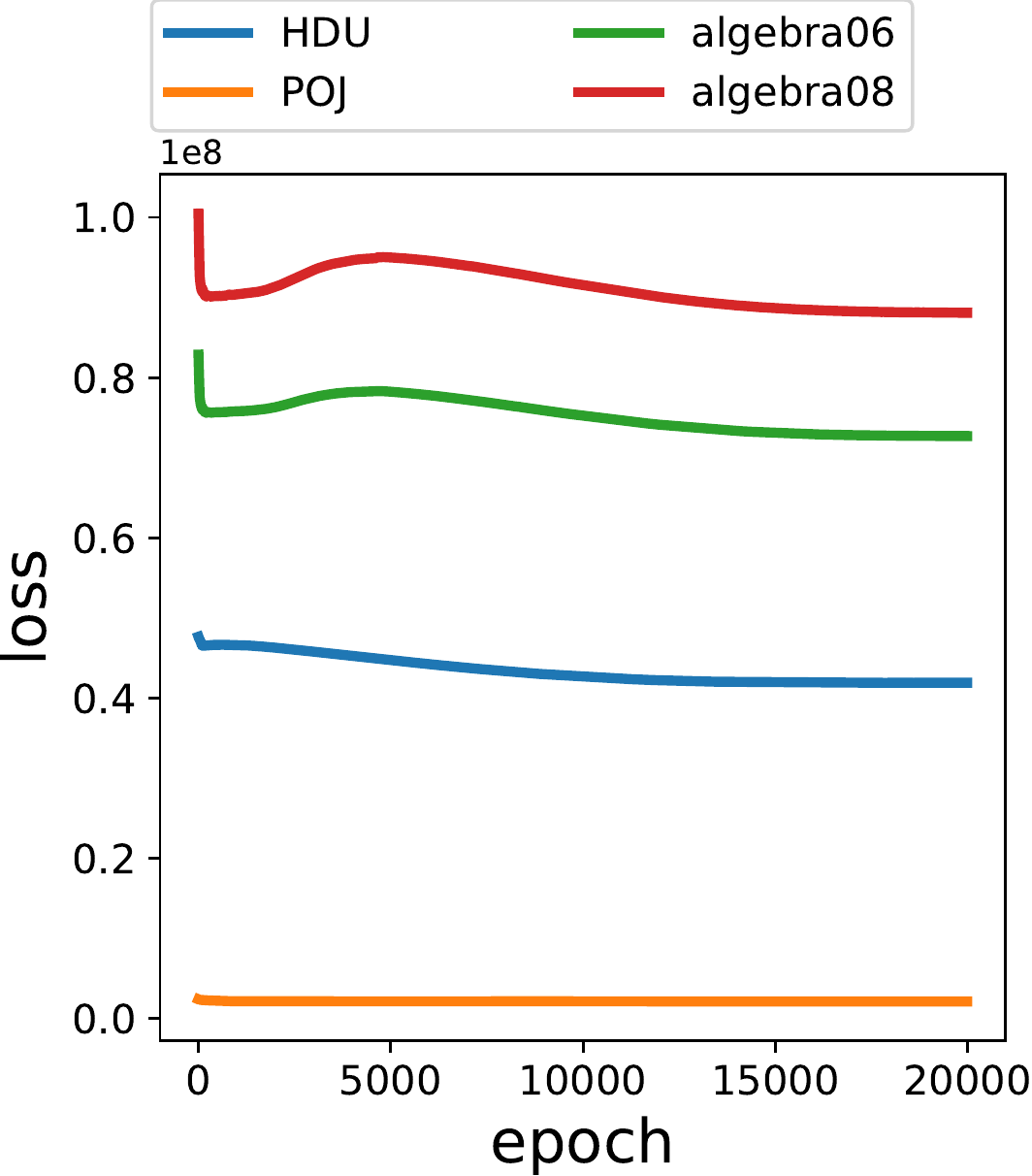}}
  \caption{\label{fig:LossHDU}\textbf{The loss values of TRACED}}
\end{figure}
\subsection{Convergence of TRACED}
\label{sec:org1227ebd}
As previously stated, TRACED is optimized by a heuristic algorithm. We prove the convergence of the heuristic algorithm we proposed with TRACED via experiments.
We have conducted experiments on four datasets, and all the experimental results show that TRACED can converge. 
As shown in Figures \ref{fig:LossHDU}, TRACED converges in the Wake and Sleep phase with four datasets. 
In the Wake phase, TRACED converges smoothly. 
In the Sleep phase, although there are some fluctuations in the convergence process, the overall trend is convergent. 
The fluctuation in the Sleep phase is caused by the unstable data generated in the Wake phase in the initial situation.
The model parameters are all obtained by random initialization. The training in the wake phase will use both the data sampled in the sleep phase and the real data, but the training data in the sleep phase is completely generated by the Wake phase. In the initial stage of training, the model parameters change too quickly, so the training data of the sleep phase is unstable and may be optimized in the wrong direction. The sleep loss will show an upward trend at the beginning. The wake phase uses the data sampled in the sleep phase and the real data, so the wake loss will not show an upward trend at the beginning.
The experimental results prove that the heuristic algorithm we proposed for the training model is a suitable and effective optimization method for learning TRACED parameters. 
\begin{figure*}[!ht]
  \centering
  \subfigure[HDU]{\includegraphics[width=0.48\textwidth]{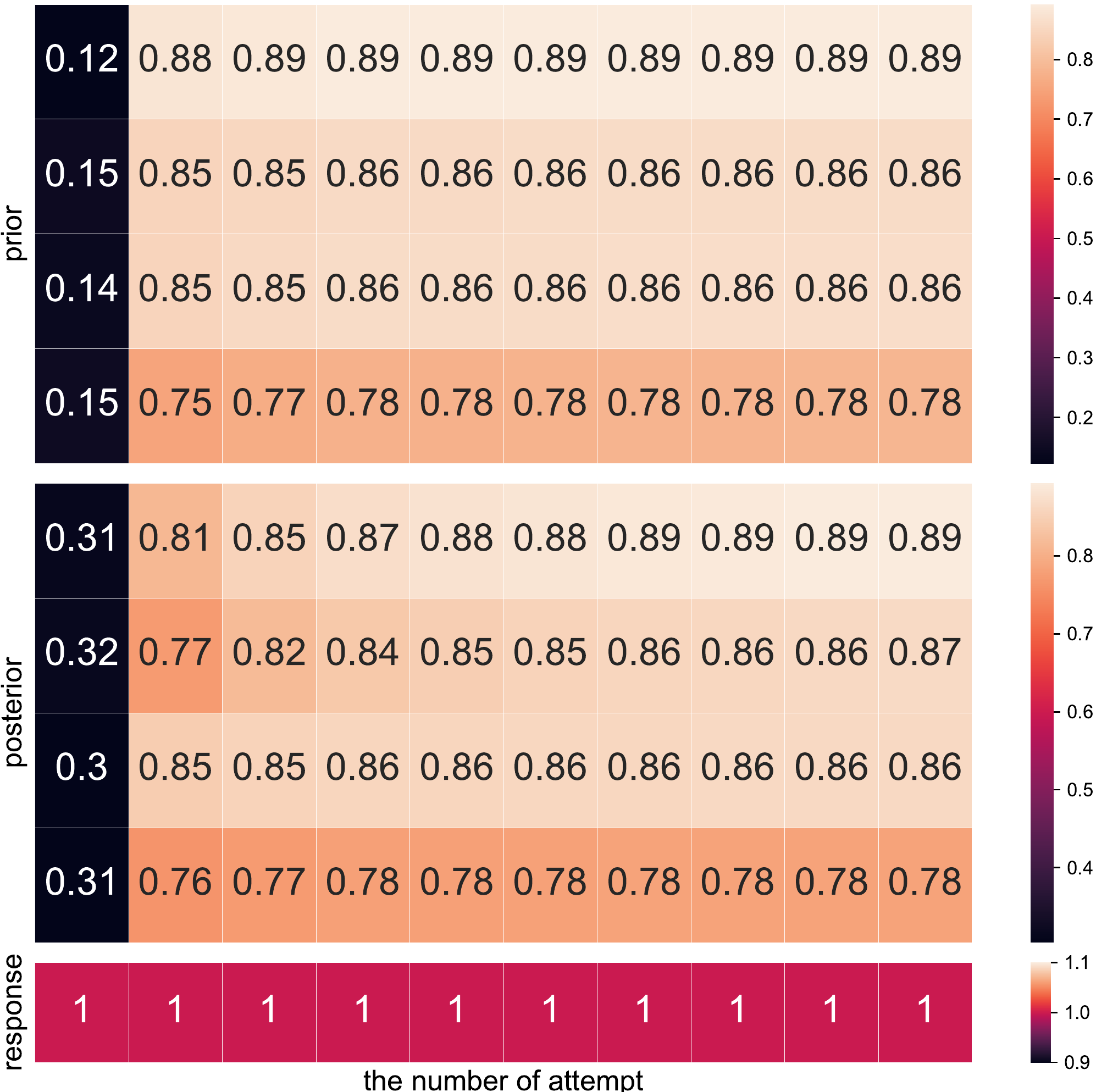}}
  \subfigure[POJ]{\includegraphics[width=0.48\textwidth]{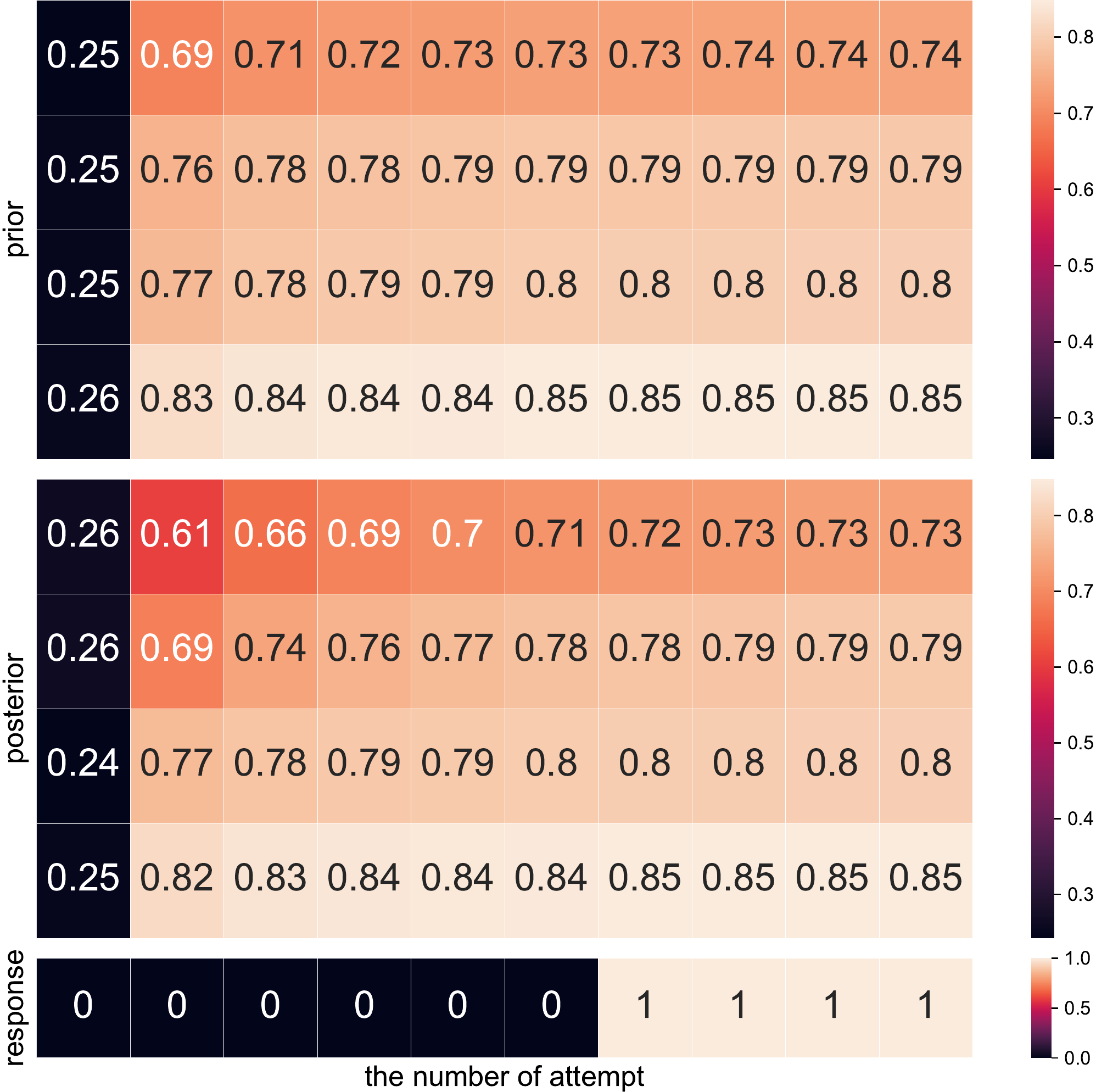}}
  \caption{\label{fig:Visualization}\textbf{The visualization of prior and posterior of TRACED on the HDU and POJ datasets}}
\end{figure*}

\subsection{Visualization of Mastery Probability}
\label{sec:orgcb40507}
Figure \ref{fig:Visualization} visualizes the prior and posterior probabilities of students' knowledge concept mastery on the HDU and POJ datasets.
The figure shows the dynamic changes in the prior and posterior probability of a student's knowledge concept mastery as students continuously practice the problem over time. 
Because the number of knowledge concepts and students is vast, we only show the dynamic change in the probability of mastering the three relevant knowledge concepts of the problem during the practice of a certain problem by a student.
With the increase in the number of exercises, the prior and posterior probability of students mastering knowledge concepts is also steadily improving.
The continuous and slow changes in the prior and posterior probabilities are consistent with our empirical understanding of the learning process. 
The calculation of the prior and posterior probabilities uses the student’s exercise problem sequence \(e_i\) and exercise time sequence \(\tau_i\), and the posterior probability can be regarded as the prior probability that is corrected based on the student’s exercise response \(r_i\). Thus, there is not much difference between the prior and posterior probability. 

The learners in the HDU and POJ datasets are the same type of learners. They are all independently programmed in the Online Judge system, so figure \ref{fig:Visualization} (a) and figure \ref{fig:Visualization} (b) can be compared horizontally. Figure \ref{fig:Visualization} (a) shows a learner who repeatedly submits the correct code. He may be adjusting the efficiency of the code. Therefore, the probability of mastering knowledge concepts increased significantly after submitting it correctly for the first time but slowly increased in subsequent submissions. Figure \ref{fig:Visualization}(b) shows that a learner is constantly practicing. With the initial six incorrect submissions, the probability of mastering the knowledge concept slowly increases. After the seventh correct submission, the probability of mastering the knowledge concept of the learner reaches its peak. Correct submission will no longer increase the probability of mastering the knowledge concept.
The above experimental results show that TRACED can provide learners and teachers with an interpretable cognitive diagnosis that changes dynamically over time.

\begin{figure}[!tbp]
\centering
\includegraphics[width=0.48\textwidth]{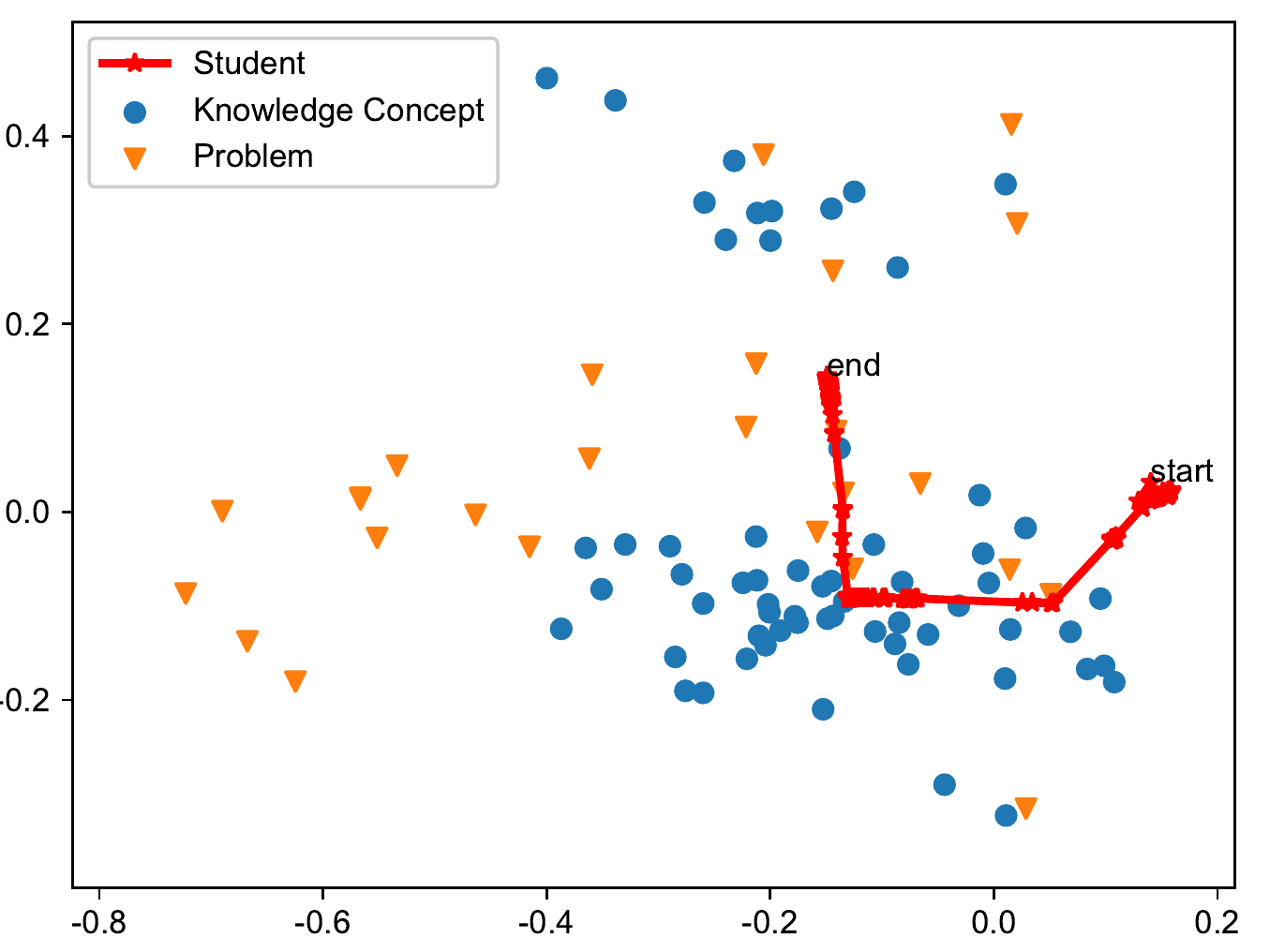}
\caption{\label{fig:ExperimentFigure1} Visualization of the learned distributed representations of students, knowledge concept and problem for the HDU dataset, where the learned representations have been reduced to 2 dimensions by means of PCA}
\end{figure}
\subsection{Visualization of Learning Trajectory}
\label{sec:orgb6ac773}
The distributed representations of the student, knowledge concepts and problems acquired using TRACED are shown in Figure \ref{fig:ExperimentFigure1}, where the acquired representations have been reduced to 2 dimensions utilizing Principal Component Analysis(PCA). 
We have drawn the learning trajectory of a learner; each point on the learning trajectory corresponds to the distributed representation of a learner after a certain exercise.
We also constructed distributed representations of problems and knowledge concepts involved in the learner's exercises.
With continuous practice, the distributed representation of the learner gradually approaches the distributed representation of knowledge concepts involved in the problem.
Besides, the result of the learner's distributed representation will cluster spontaneously, and there is a certain distance between each cluster. The result of the learner's distributed representation within the class shows a straight line. Because in addition to learning by practicing on the online education platform, students can also learn in other ways (offline teachers teach, read paper textbooks, and soon on.). We cannot collect the data generated by these learning processes. Thus, there are significant jumps between two certain clusters on the student's learning trajectory.
The reason why the learning trajectory of the learner in the class shows a straight line is that the results of the distributed representation of the learner in the class are generally produced by a phased exercise. The knowledge concepts involved in the phase exercises are relatively similar, so the learner's embedding result shifts toward the specified direction.
By using the learning trajectory of students, teachers can quickly understand the learning habits and learning process of students and then can customize personalized learning plans for students conveniently and quickly.

%% file: src/6-Conclusion.tex
\section{Conclusions}
\label{sec:org1dcb972}
In this paper, we proposed an inTerpretable pRobAbilistiC gEnerative moDel (TRACED), which can track numerous students' knowledge concept mastery probabilities over time. 
To better model students' learning process, we adopted the learning and forgetting curves as priors to capture the dynamic changes in students' knowledge concept mastery over time. 
Second, we designed a logarithmic linear model with three interactive strategies to model students' exercise responses by considering the interactions among knowledge concepts, problems, and students.
To solve \emph{explain away} problem, we design LSTM-based networks to approximate the posterior distribution and propose a heuristic algorithm to learn model parameters. 
Lastly, We conduct experiments with four real-world datasets in three knowledge-driven tasks. 
The experimental results for predicting students' future performance demonstrate the effectiveness of TRACED as a knowledge tracing model.
The experimental results for predicting the relationship between knowledge concepts and the relationship between knowledge concepts and problems demonstrate the effectiveness of the distributed representation of knowledge concepts and problems.
Besides, We conduct several case studies. The case studies show that TRACED exhibits an excellent interpretable capability and has the potential for personalized automatic feedback in a real-world educational environment. 

\section{Acknowledgments}
The work described in this paper was supported by National Natural Science Foundation of China (Nos.62272093, 62137001, U1811261, 61902055). 

%% file: src/7-Author.tex
\begin{biography}{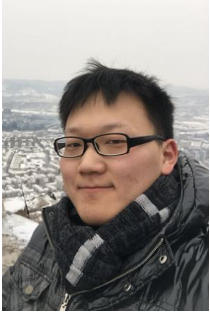}
Hengyu Liu received a B.S. degree in computer science from Northeastern University, China, in 2017. He is currently a Ph.D. student studying computer software and theory at Northeastern University, China. His research interests include machine learning, deep learning, graph model, cognitive diagnosis and knowledge tracing.
\end{biography}
\begin{biography}{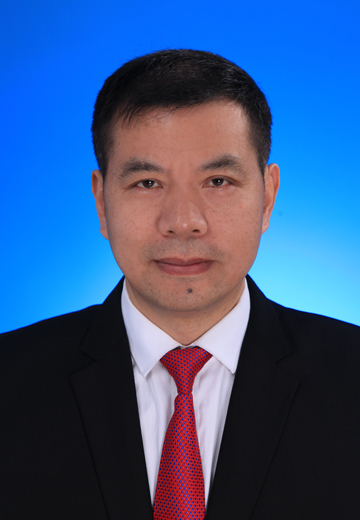}
Tiancheng Zhang received a Ph.D degree in computer software and theory from Northeastern University (NEU) of China. He is currently an associate professor in the School of Computer Science and Engineering at NEU. His research interests include big data analysis, spatiotemporal data management, and deep learning.
\end{biography}
\begin{biography}{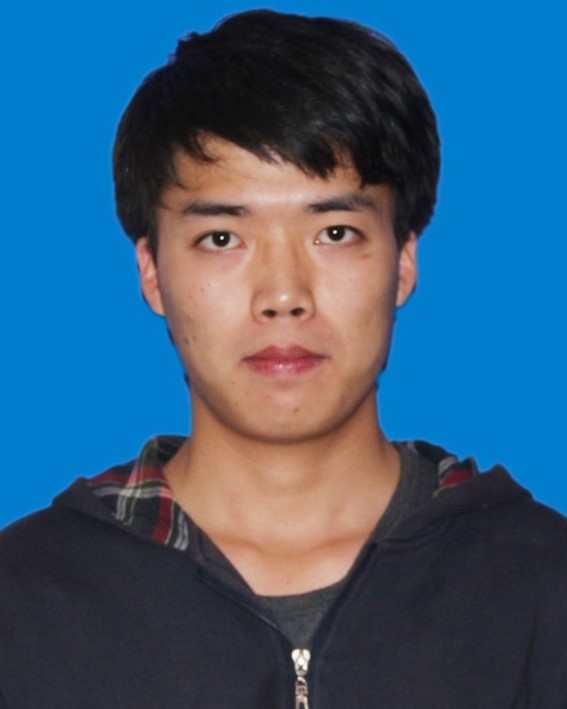}
Fan Li received a B.E. degree in computer science from Qinghai University in 2019. He is currently a Master's student at the School of Computer Science And Engineering, Northeastern University, China. His research interest lies in artificial intelligence in education. 
\end{biography}
\begin{biography}{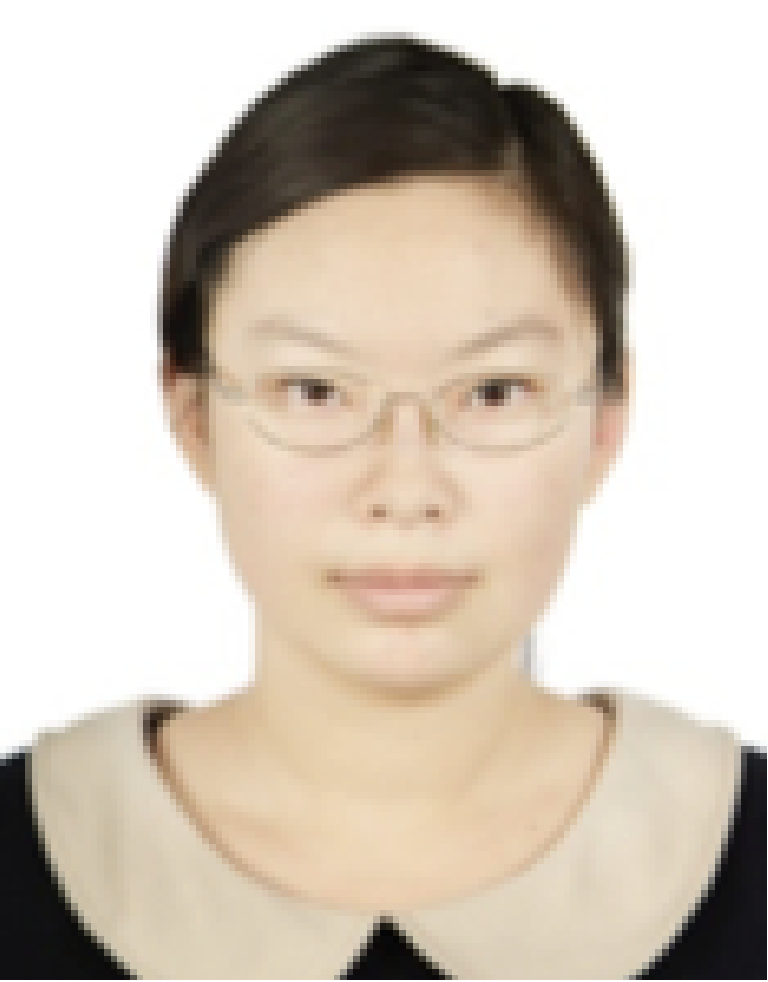}
Minghe Yu was born in Shenyang, Liaoning, China, in 1989. She received the B.S. degree in computer science and technology from Northeastern University, Shenyang, in 2012, and the Ph.D. degree in computer science and technology from Tsinghua University, Beijing, China, in 2018. Since 2018, she has been a lecturer with the Software College, Northeastern University. Her research interests include big data, information retrieval, and data mining.
\end{biography}
\begin{biography} {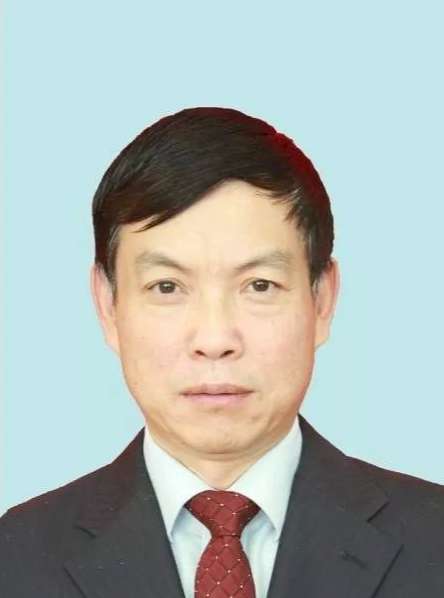}
Ge Yu received Ph.D. degree in computer science from Kyushu University, Japan, in 1996. He is currently a professor and a Ph.D. Supervisor at Northeastern University, China. His research interests include distributed and parallel databases, OLAP and data warehousing, data integration, and graph data management. He is a member of ACM and a Fellow of the China Computer Federation (CCF).
\end{biography}

%% file: main.bbl
\begin{thebibliography}{10}

\bibitem{DBLP:conf/kdd/LiuTLZCMW19}
Liu Q, Tong S, Liu C, Zhao H, Chen E, Ma~H, Wang S.
\newblock Exploiting cognitive structure for adaptive learning.
\newblock In: Proceedings of the 25th ACM SIGKDD International Conference on
  Knowledge Discovery \& Data Mining, KDD '19.
\newblock 2019,  627–635

\bibitem{DBLP:conf/ijcai/WuLLCSCH15}
Wu~R, Liu Q, Liu Y, Chen E, Su~Y, Chen Z, Hu~G.
\newblock Cognitive modelling for predicting examinee performance.
\newblock In: {IJCAI}.
\newblock 2015,  1017--1024

\bibitem{DBLP:conf/edm/AiCGZWFW19}
Ai~F, Chen Y, Guo Y, Zhao Y, Wang Z, Fu~G, Wang G.
\newblock Concept-aware deep knowledge tracing and exercise recommendation in
  an online learning system.
\newblock In: The 12th International Conference on Educational Data Mining,
  EDM'2019.
\newblock 2019

\bibitem{DBLP:conf/nips/PiechBHGSGS15}
Piech C, Bassen J, Huang J, Ganguli S, Sahami M, Guibas L~J.
\newblock Deep knowledge tracing.
\newblock In: {NIPS}.
\newblock 2015,  505--513

\bibitem{DBLP:conf/cikm/ChenLHWCWSH17}
Chen Y, Liu Q, Huang Z, Wu~L, Chen E, Wu~R, Su~Y, Hu~G.
\newblock Tracking knowledge proficiency of students with educational priors.
\newblock In: Proceedings of the 2017 ACM on Conference on Information and
  Knowledge Management.
\newblock 2017,  989--998

\bibitem{Sun2022}
Sun S, Hu~X, Bu~C, Liu F, Zhang Y, Luo W.
\newblock Genetic algorithm for bayesian knowledge tracing: {A} practical
  application.
\newblock In: Tan Y, Shi Y, Niu B, eds, Advances in Swarm Intelligence - 13th
  International Conference, {ICSI} 2022, Xi'an, China, July 15-19, 2022,
  Proceedings, Part {I}.
\newblock 2022,  282--293

\bibitem{Wong2021}
Wong T, Zou D, Cheng G, Tang J~K~T, Cai Y, Wang F~L.
\newblock Enhancing skill prediction through generalising bayesian knowledge
  tracing.
\newblock Int. J. Mob. Learn. Organisation, 2021, 15(4): 358--373

\bibitem{Liu2022}
Liu F, Hu~X, Bu~C, Yu~K.
\newblock Fuzzy bayesian knowledge tracing.
\newblock {IEEE} Trans. Fuzzy Syst., 2022, 30(7): 2412--2425

\bibitem{Zhang2018}
Zhang K, Yao Y.
\newblock A three learning states bayesian knowledge tracing model.
\newblock Knowl. Based Syst., 2018, 148: 189--201

\bibitem{204911}
{Wellman} M~P, {Henrion} M.
\newblock Explaining 'explaining away'.
\newblock IEEE Transactions on Pattern Analysis and Machine Intelligence, 1993,
  15(3): 287--292

\bibitem{van2013handbook}
Linden v.~d W~J, Hambleton R~K.
\newblock Handbook of modern item response theory.
\newblock Springer Science \& Business Media, 2013

\bibitem{cen2006learning}
Cen H, Koedinger K, Junker B.
\newblock Learning factors analysis--a general method for cognitive model
  evaluation and improvement.
\newblock In: International Conference on Intelligent Tutoring Systems.
\newblock 2006,  164--175

\bibitem{pavlik2009performance}
Pavlik~Jr P~I, Cen H, Koedinger K~R.
\newblock Performance factors analysis--a new alternative to knowledge tracing.
\newblock Online Submission, 2009

\bibitem{DBLP:conf/aaai/VieK19}
Vie J, Kashima H.
\newblock Knowledge tracing machines: Factorization machines for knowledge
  tracing.
\newblock In: {AAAI}.
\newblock 2019,  750--757

\bibitem{DBLP:conf/edm/ChoffinPBV19}
Choffin B, Popineau F, Bourda Y, Vie J.
\newblock {DAS3H:} modeling student learning and forgetting for optimally
  scheduling distributed practice of skills.
\newblock In: {EDM}.
\newblock 2019

\bibitem{corbett1994knowledge}
Corbett A~T, Anderson J~R.
\newblock Knowledge tracing: Modeling the acquisition of procedural knowledge.
\newblock User modeling and user-adapted interaction, 1994, 4(4): 253--278

\bibitem{Gorgun2022}
Gorgun G, Bulut O.
\newblock Considering disengaged responses in bayesian and deep knowledge
  tracing.
\newblock In: Rodrigo M~M~T, Matsuda N, Cristea A~I, Dimitrova V, eds,
  Artificial Intelligence in Education. Posters and Late Breaking Results,
  Workshops and Tutorials, Industry and Innovation Tracks, Practitioners' and
  Doctoral Consortium - 23rd International Conference, {AIED} 2022, Durham, UK,
  July 27-31, 2022, Proceedings, Part {II}.
\newblock 2022,  591--594

\bibitem{de2009dina}
De~La~Torre J.
\newblock Dina model and parameter estimation: A didactic.
\newblock Journal of educational and behavioral statistics, 2009, 34(1):
  115--130

\bibitem{liu2020tracking}
Liu H, Zhang T, Li~F, Gu~Y, Yu~G.
\newblock Tracking knowledge structures and proficiencies of students with
  learning transfer.
\newblock IEEE Access, 2020, 9: 55413--55421

\bibitem{ausubel1968educational}
Ausubel D~P, Novak J~D, Hanesian H, others .
\newblock Educational psychology: A cognitive view. volume~6.
\newblock holt, rinehart and Winston New York, 1968

\bibitem{DBLP:conf/aaai/SuLLHYCDWH18}
Su~Y, Liu Q, Liu Q, Huang Z, Yin Y, Chen E, Ding C~H~Q, Wei S, Hu~G.
\newblock Exercise-enhanced sequential modeling for student performance
  prediction.
\newblock In: {AAAI}.
\newblock 2018,  2435--2443

\bibitem{huang2019ekt}
Huang Z, Yin Y, Chen E, Xiong H, Su~Y, Hu~G, others .
\newblock Ekt: Exercise-aware knowledge tracing for student performance
  prediction.
\newblock IEEE Transactions on Knowledge and Data Engineering, 2019

\bibitem{lu2020towards}
Lu~Y, Wang D, Meng Q, Chen P.
\newblock Towards interpretable deep learning models for knowledge tracing.
\newblock In: International Conference on Artificial Intelligence in Education.
\newblock 2020,  185--190

\bibitem{DBLP:conf/www/ZhangSKY17}
Zhang J, Shi X, King I, Yeung D.
\newblock Dynamic key-value memory networks for knowledge tracing.
\newblock In: {WWW}.
\newblock 2017,  765--774

\bibitem{sun2019muti}
Sun X, Zhao X, Ma~Y, Yuan X, He~F, Feng J.
\newblock Muti-behavior features based knowledge tracking using decision tree
  improved dkvmn.
\newblock In: Proceedings of the ACM Turing Celebration Conference-China.
\newblock 2019,  1--6

\bibitem{DBLP:conf/edm/PandeyK19}
Pandey S, Karypis G.
\newblock A self attentive model for knowledge tracing.
\newblock In: {EDM}.
\newblock 2019

\bibitem{DBLP:conf/cikm/PandeyS20}
Pandey S, Srivastava J.
\newblock {RKT:} relation-aware self-attention for knowledge tracing.
\newblock In: {CIKM}.
\newblock 2020,  1205--1214

\bibitem{ghosh2020context}
Ghosh A, Heffernan N, Lan A~S.
\newblock Context-aware attentive knowledge tracing.
\newblock In: Proceedings of the 26th ACM SIGKDD International Conference on
  Knowledge Discovery \& Data Mining.
\newblock 2020,  2330--2339

\bibitem{zhu2020learning}
Zhu J, Yu~W, Zheng Z, Huang C, Tang Y, Fung G~P~C.
\newblock Learning from interpretable analysis: Attention-based knowledge
  tracing.
\newblock In: International Conference on Artificial Intelligence in Education.
\newblock 2020,  364--368

\bibitem{yu2022contextkt}
Yu~M, Li~F, Liu H, Zhang T, Yu~G.
\newblock Contextkt: A context-based method for knowledge tracing.
\newblock Applied Sciences, 2022, 12(17): 8822

\bibitem{DBLP:conf/kdd/0001HL20}
Ghosh A, Heffernan N~T, Lan A~S.
\newblock Context-aware attentive knowledge tracing.
\newblock In: {KDD}.
\newblock 2020,  2330--2339

\bibitem{anzanello2011learning}
Anzanello M~J, Fogliatto F~S.
\newblock Learning curve models and applications: Literature review and
  research directions.
\newblock International Journal of Industrial Ergonomics, 2011, 41(5): 573--583

\bibitem{von2007understanding}
Von~Foerster H.
\newblock Understanding understanding: Essays on cybernetics and cognition.
\newblock Springer Science \& Business Media, 2007

\bibitem{rabiner1989tutorial}
Rabiner L~R.
\newblock A tutorial on hidden markov models and selected applications in
  speech recognition.
\newblock Proceedings of the IEEE, 1989, 77(2): 257--286

\bibitem{hinton1995wake}
Hinton G~E, Dayan P, Frey B~J, Neal R~M.
\newblock The" wake-sleep" algorithm for unsupervised neural networks.
\newblock Science, 1995, 268(5214): 1158--1161

\bibitem{fogarty2005case}
Fogarty J, Baker R~S, Hudson S~E.
\newblock Case studies in the use of roc curve analysis for sensor-based
  estimates in human computer interaction.
\newblock In: Proceedings of Graphics Interface 2005.
\newblock 2005,  129--136

\bibitem{liu2022learning}
Liu H, Fu~Q, Du~L, Zhang T, Yu~G, Han S, Zhang D.
\newblock Learning rate perturbation: A generic plugin of learning rate
  schedule towards flatter local minima.
\newblock In: Proceedings of the 31st ACM International Conference on
  Information \& Knowledge Management.
\newblock 2022,  4234--4238

\bibitem{embretson2013item}
Embretson S~E, Reise S~P.
\newblock Item response theory.
\newblock Psychology Press, 2013

\bibitem{lindsey2014improving}
Lindsey R~V, Shroyer J~D, Pashler H, Mozer M~C.
\newblock Improving students’ long-term knowledge retention through
  personalized review.
\newblock Psychological science, 2014, 25(3): 639--647

\bibitem{mozer2016predicting}
Mozer M~C, Lindsey R~V.
\newblock Predicting and improving memory retention.
\newblock Big data in cognitive science, 2016, ~34

\bibitem{pmlr-v9-glorot10a}
Glorot X, Bengio Y.
\newblock Understanding the difficulty of training deep feedforward neural
  networks.
\newblock In: Teh Y~W, Titterington M, eds, Proceedings of the Thirteenth
  International Conference on Artificial Intelligence and Statistics.
\newblock 13--15 May 2010,  249--256

\bibitem{mikolov2013distributed}
Mikolov T, Sutskever I, Chen K, Corrado G~S, Dean J.
\newblock Distributed representations of words and phrases and their
  compositionality.
\newblock In: Advances in neural information processing systems.
\newblock 2013,  3111--3119

\end{thebibliography}
